\documentclass[10pt,twocolumn,letterpaper]{article}

\usepackage{cvpr}              

\usepackage[utf8]{inputenc}
\usepackage[danish,english]{babel}
\usepackage{csquotes}

\usepackage[T1]{fontenc}
\usepackage{esvect} 

\usepackage{listings}

\usepackage{graphicx}
\graphicspath{./figures/}
\usepackage{booktabs}
\usepackage{framed}
\usepackage{pgfplots}
\usetikzlibrary{patterns}
\usepgfplotslibrary{statistics}
\pgfplotsset{compat=1.18}

\usepackage{array}
\newcolumntype{P}[1]{>{\centering\arraybackslash}p{#1}}

\usepackage{svg}
\usepackage{amsmath}
\usepackage{amssymb}
\usepackage[framed,amsmath,thmmarks]{ntheorem}

\usepackage{fancyhdr}

\usepackage{calc}

\usepackage{pdflscape}
\usepackage{rotating}

\usepackage{float}
\usepackage{tabu}
\usepackage{gensymb}
\usepackage{textcomp}
\usepackage{multirow}
\usepackage{enumitem}
\usepackage[bottom]{footmisc}

\definecolor{cvprblue}{rgb}{0.21,0.49,0.74}
\usepackage[pagebackref,breaklinks,colorlinks,allcolors=cvprblue]{hyperref}

\def\paperID{42} 
\def\confName{CVPR}
\def\confYear{2025}

\title{Point Cloud Segmentation of Agricultural Vehicles using 3D Gaussian Splatting}

\usepackage{subcaption}
\usepackage{comment}
\usepackage{lipsum}
\usepackage{graphicx}
\definecolor{ground}{rgb}{0.050383, 0.029803, 0.527975}
\definecolor{tractor}{rgb}{0.610667, 0.090204, 0.619951}
\definecolor{combine}{rgb}{0.928329, 0.472975, 0.326067}

\definecolor{greenbbx}{rgb}{0.00392156862745098, 0.6980392156862745, 0}
\definecolor{redbbx}{rgb}{1,0,0}

\usepackage{xcolor}

\DeclareRobustCommand{\legendsquare}[1]{%
  \textcolor{#1}{\rule{1ex}{1ex}}%
}

\author{
Alfred T. Christiansen$^{1,\dagger}$ \quad Andreas H. Højrup$^{1,\dagger}$ \quad Morten K. Stephansen$^{1,\dagger}$\\ Md Ibtihaj A. Sakib$^{1}$ \quad Taman S. Poojary$^{1}$\\ Filip Slezák$^{2,3}$ \quad Morten S. Laursen$^{3}$ \quad Thomas B. Moeslund$^{2,4}$ \quad Joakim B. Haurum$^{2,4}$\\
$^{1}$Department of Electronic Systems, Aalborg University\quad $^{3}$AGCO A/S, Denmark\\ $^{2}$Visual Analysis \& Perception Lab, Aalborg University\quad $^{4}$Pioneer Centre for AI, Denmark\\
}

\begin{document}

\newcommand{\figfirstpagefigure}{
\vspace{-2em}
\begin{center}
\captionsetup{type=figure}
\includegraphics[width=\textwidth]{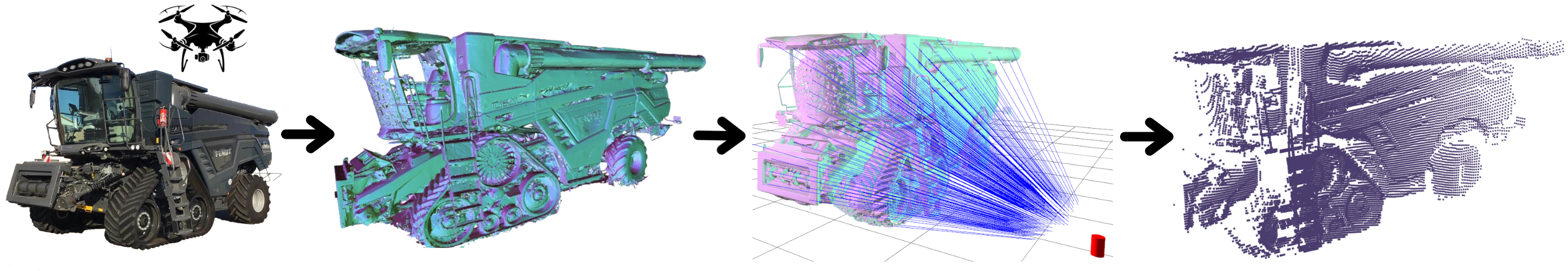}
\vspace{-2em}
\captionof{figure}{
Illustration shows the pipeline created for generating the synthetic point cloud dataset. Step (1) shows the process of gathering images of a combine harvester. Step (2) shows the combine harvester mesh, extracted using GOF. Step (3) shows an example of a combine harvester mesh employed in the Gazebo LiDAR simulation environment, where the LiDAR sensor is visualized as the red cylinder with rays depicting the LiDAR scans. Step (4) shows an example of a final, synthetically generated, point cloud of a combine harvester.}
\label{fig:teaser}
\end{center}
}
\twocolumn[{
\maketitle
\figfirstpagefigure
}]


\begin{abstract}
Training neural networks for tasks such as 3D point cloud semantic segmentation demands extensive datasets, yet obtaining and annotating real-world point clouds is costly and labor-intensive. This work aims to introduce a novel pipeline for generating realistic synthetic data, by leveraging 3D Gaussian Splatting (3DGS) and Gaussian Opacity Fields (GOF) to generate 3D assets of multiple different agricultural vehicles instead of using generic models. These assets are placed in a simulated environment, where the point clouds are generated using a simulated LiDAR. This is a flexible approach that allows changing the LiDAR specifications without incurring additional costs. We evaluated the impact of synthetic data on segmentation models such as PointNet++, Point Transformer V3, and OACNN, by training and validating the models only on synthetic data. Remarkably, the PTv3 model had an mIoU of 91.35\%, a noteworthy result given that the model had neither been trained nor validated on any real data. Further studies even suggested that in certain scenarios the models trained only on synthetically generated data performed better than models trained on real-world data. Finally, experiments demonstrated that the models can generalize across semantic classes, enabling accurate predictions on mesh models they were never trained on.
\end{abstract}

\section{Introduction} \label{sec:introduction}
The demand for automation in the transportation sector for self driving vehicles has, in recent years, fueled research for having machines understand 3D environments \cite{Sun_2020_CVPR}. As a format to capture 3D information, point clouds are increasingly popular due to their efficiency and direct representation of LiDAR data. 
Commonly, for the models to understand the composition of the environment represented by the point cloud, semantic segmentation is used as it provides per-point class-wise information, thus getting an understanding of the complete environment \cite{qi2017pointnetdeeplearningpoint, qi2017pointnetdeephierarchicalfeature, wang2019dynamic, peng2024oacnnsomniadaptivesparsecnns, zhao2021point, wu2022point, wu2024pointtransformerv3simpler, li2018pointcnn}. An example of this is autonomous driving, where point cloud semantic segmentation is used to identify vehicles, road signs, pedestrians, etc. \cite{Sun_2020_CVPR}.

However, obtaining and annotating point clouds is a well-known tedious and expensive task, making the available training resources sparse compared to the training datasets for image recognition. This has led a wave of research into both using the point clouds more efficiently by augmenting them \cite{chen2020pointmixup, li2020pointaugment}, and methods for generating new, annotated point clouds synthetically using simulated environments relying on existing 3D assets \cite{wang2019automatic, yue2018lidar, ma2020semantic}. The issue with using existing 3D assets arises when models need training for niche tasks with a sparse amount of available 3D assets. As a method to mitigate this issue, we propose the use of 3D Gaussian splatting (3DGS) to generate 3D assets that can be used to generate synthetic point cloud datasets. 3DGS is used due to the ease of only needing a digital camera for capturing useful data, and a few hours of GPU time to get a mesh. 
In this project, the method is tested as a use case to generate and train point cloud semantic segmentation models in a self driving tractor scenario, with synthetic data acquired using the pipeline shown in Figure~\ref{fig:teaser}. This is a motivating domain for the proposed pipeline, as the existing 3D data is sparse and the meshes needed would be hard to model from scratch. 
As a proof of concept, the class count is reduced to: tractor, combine harvester, and other, as these vehicles often carry out collaborative tasks. Our key contributions are as follows:

\begin{itemize}
    \item We present a novel way of generating high-quality LiDAR data from 3DGS.
    \item We demonstrate that by training a model purely on synthetic data it is possible to achieve high performance on real world data.
    \item We show that our method generalizes to objects not seen during training, highlighting its robustness and adaptability.
    \item Our method allows seamless modification of LiDAR-specific settings and the choice of mesh models used in the simulation, enabling greater adaptability to different sensing conditions and environments. 
\end{itemize}
\section{Related work}
\subsection{Point Cloud Semantic Segmentation}
%
Segmentation of 3D points is a common problem in the field of 3D computer vision. Many methods have been proposed to tackle the challenge of segmenting the unstructured and irregular-spaced point clouds \cite{qi2017pointnetdeeplearningpoint, qi2017pointnetdeephierarchicalfeature, wang2019dynamic, peng2024oacnnsomniadaptivesparsecnns, zhao2021point, wu2022point, wu2024pointtransformerv3simpler, li2018pointcnn}. These methods can be categorized into two categories: point-based models and voxel-based models \cite{peng2024oacnnsomniadaptivesparsecnns}. Point-based methods directly manipulate the unstructured point clouds using point-wise operations. One such method is PointNet \cite{qi2017pointnetdeeplearningpoint}, which introduced a novel deep learning architecture used for point cloud semantic segmentation. However, the network cannot capture local structures due to the lack of neighborhood processing, limiting its ability to recognize fine-grained patterns and generalize to complex scenes. This was improved in PointNet++ \cite{qi2017pointnetdeephierarchicalfeature} by partitioning the point set into local overlapping regions, and then recursively applying PointNet to obtain the local features. Dynamic Graph CNN, proposed by Wang \etal \cite{wang2019dynamic}, is another method which expands upon the PointNet architecture, this time by implementing a new convolutional operation, EdgeConv, which allows the extraction of the local neighborhood features for each point. Another prominent point-based method is the transformer-based models, which leverages the attention mechanism \cite{vaswani2017attention} that has gained prominence across various domains, including point cloud segmentation. The Point Transformer V3 (PTv3) \cite{wu2024pointtransformerv3simpler} advances point cloud segmentation by emphasizing scalability and efficiency. Unlike its predecessors, PTv1 \cite{zhao2021point} and PTv2 \cite{wu2022point}, which introduced local attention mechanisms, grouped vector attention, improved position encoding, and partition-based pooling, PTv3 focuses on the benefits of scaling rather than architectural complexity. It replaces the K-Nearest Neighbors (KNN) query with serialized patterns like Z-order and Hilbert curves, significantly reducing computational overhead, as KNN accounted for 28\% of PTv2’s forward time. Additionally, PTv3 refines the positional encoding, further enhancing the model's efficiency and ability to handle large-scale point clouds.

Voxel-based methods convert the point clouds into a voxel grid during data pre-processing. A recent paper by Peng \etal \cite{peng2024oacnnsomniadaptivesparsecnns}, introduced an Omni-Adaptive CNN which addressed the common problem of adaptivity for previous voxel-based methods \cite{peng2024oacnnsomniadaptivesparsecnns}. They created an Adaptive Relation Convolution, and a corresponding adaptive aggregator, which dynamically adjusts the receptive field, allowing the model to focus more on parts with many defining features.

\subsection{Synthetic data generation}
\label{subsec:synth_data_related}
Multiple methods have been proposed to generate synthetic annotated point cloud datasets \cite{Xiao2021SynLiDAR, ma2020semantic, wang2019automatic, yue2018lidar, takmaz20233dhumansegemnataion, chen2020pointmixup, li2020pointaugment}. These methods can be categorized into using augmented real point clouds or using simulated environments. 
A recent method that uses simulated point clouds that are then further processed by an augmentation process is suggested by Xiao \etal \cite{Xiao2021SynLiDAR}, proposing the use of the Unreal Engine to create a synthetic LiDAR segmentation dataset where they used an adversarial network to transform the synthetic point clouds that are acquired from the simulated environment, into point clouds that are closer to what would be sampled in reality. 

Chen \etal \cite{chen2020pointmixup} proposes to linearly interpolate between point clouds of the same classes to generate new training data. 
Another augmentation method by Li \etal \cite{li2020pointaugment} propose an augmentation neural network trained together with a classifier neural network in an adversarial manner. However, this method has not been tested on point cloud segmentation tasks.

Focusing on simulated environments, Ma \etal \cite{ma2020semantic} uses building models to create synthetic annotated point clouds for training segmentation tasks. The point cloud sampling method used in the study resulted in uniformly sampled point clouds, whereas real point clouds would have non-uniform sparsity. This, according to the paper, is something that could be improved upon, as synthetic data should be similar to the real data.

Another simulated approach by Wang \etal \cite{wang2019automatic} uses a simulated city environment made with the CARLA simulation tool \cite{dosovitskiy2017carla}, where the LiDAR is simulated with ray casting to get a synthetic segmented LiDAR point cloud dataset. Similarly, Yue \etal \cite{yue2018lidar} uses the video game Grand Theft Auto V to simulate an outdoor driving environment. Using plugins, a synthetic dataset is generated from the simulation. 
An approach that focuses on human body part segmentation proposed by Takamz \etal \cite{takmaz20233dhumansegemnataion}, used mesh scenes from the ScanNet dataset\cite{dai2017scannet} with human models, where the poses of the humans have been generated. These scenes are then sampled with a simulated depth camera to obtain annotated point clouds. This method relies on having a pre-existing dataset of scenes, and a model \cite{zhang2020place} to generate synthetic humans in the scene. Due to this, it is ill suited for making datasets of niche tasks. 

\subsection{Novel View Synthesis}
Novel view synthesis aims to generate an unseen view of a scene from an arbitrary viewpoint. It can, however, also be used to capture the geometry of a scene. Mildenhall \etal \cite{mildenhall2020nerf} introduced Neural Radiance Field (NeRF), which uses a multi-layer perceptron (MLP) to represent a 3D scene, including view-dependent reflections, colors, and geometry. Although later enhancements in NeRF have improved the rendering speed \cite{hedman2021baking, yariv2023bakedsdf, reiser2023merf}, and anti-aliasing \cite{barron2022mip, barron2023zip}, they are still implicit, as the scene is encoded in the weights of the model. This results in extra long inference times that can be reduced only by lowering the quality of the visuals. Another method used to synthesize novel views is 3D Gaussian Splatting, introduced by Kerbl \etal \cite{kerbl20233dgaussiansplattingrealtime}, which uses 3D Gaussians to represent a scene explicitly, thus allowing real-time rendering, editable scenes, and a more accurate extraction of geometry. Furthermore, in recent years NeRF and 3DGS methods have been used to generate synthetic data, though so far limited to dense image tasks such as stereo vision and optical flow \cite{ADFactory, FilipCVPRW, NerfStereo, SelfEvlovling3DGS}.

\subsection{Mesh Extraction}
Photorealistic rendering through 3DGS has shown remarkable efficiency compared to NeRFs, however, generating accurate geometric reconstructions from these scene representations, remains a challenging problem. This challenge arises from the inherently disconnected nature of individual Gaussian primitives, and the complexity of aligning these Gaussians with continuous surfaces for reconstruction. \cite{guedon2024sugar, huang20242d, yu2024gaussian}

SuGaR, introduced by Gu\'edon \etal \cite{guedon2024sugar}, regularizes the 3D gaussians to align with surfaces, allowing them to compute the surface normals. Using the regularized 3D gaussians and the computed normals, a Poisson surface reconstruction is employed to generate the mesh. 
Huang \etal \cite{huang20242d} proposes 2DGS, which, instead of the original 3D gaussians splats, uses 2D gaussians to recreate the radiance field. The meshes are then reconstructed through depth maps of the radiance field using Truncated Signed Distance Function. 

Yu \etal \cite{yu2024gaussian} introduced Gaussian Opacity Fields (GOF), a state-of-the-art technique that directly extracts surface normals from the 3D Gaussian representations, without requiring prior regularization or conversion to 2D. This method mitigates the inevitable data loss that happens when regularizing or reducing the dimension of the radiance field, enhancing the quality in more detailed parts. Using the extracted normals, the final mesh extraction is done by utilizing tetrahedral grids and the Marching Tetrahedra algorithm.
\section{Methods}

\subsection{Data Acquisition \& Processing Pipeline}
To evaluate the influence of synthetic data on a point cloud semantic segmentation model, a \textit{baseline} model is trained using only 'real' data acquired with an Ouster OS0 LiDAR \cite{OusterOS0}. A baseline model will be trained for each model tested in the paper. The effect of using synthetic data will be established based on a comparison between the baseline models, and models trained on both synthetic and real data. Figure \ref{fig:pipeline} shows how data is acquired for the datasets, where the training and validation data is comprised of synthetic data. It also includes the pipeline for obtaining synthetic data, which begins with capturing 700+ images of each vehicle using a drone. These images are then processed to extract 3D meshes using GOF \cite{yu2024gaussian}. Finally, the meshes are imported into Gazebo \cite{koenig2004gazebo}, where a simulated LiDAR sensor is used to generate the synthetic data.


\begin{figure}[t]
  \centering
  \includegraphics[width=0.85\linewidth]{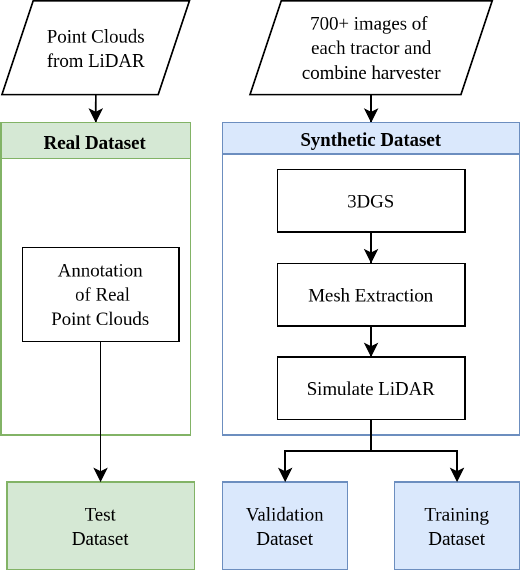}
  \caption{Illustration of the pipeline used to obtain synthetic datasets.}
\end{figure}

\subsection{LiDAR captured dataset} \label{subsec:lidar_captured_dataset}
\begin{figure}
    \centering
    \includegraphics[width=0.49\linewidth]{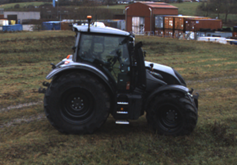}
    \includegraphics[width=0.49\linewidth]{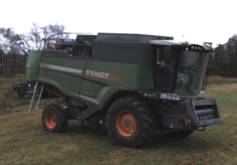}
    \caption{Drone-captured frames of a tractor (left) and a combine harvester (right) used in the 3D Gaussian Splatting mesh generation process.}
    \label{fig:tractor_and_combine_images}
\end{figure}

The real dataset, which is split into training, validation and testing, is created by capturing LiDAR data from the real world. The data capture process is performed by driving around in a tractor, with the Ouster OS0-128 LiDAR sensor mounted on top, in a common agricultural scenario with multiple tractors and combine harvesters. Figure \ref{fig:tractor_and_combine_images} presents an image of one of the tractors and an image of one of the combine harvesters. The data is collected in sequences of driving around for one minute, with the OS0-128 LiDAR sensor sampling point clouds at 10 Hz, yielding 600 point clouds per sequence. In total, 15 sequences from six different configurations have been acquired and annotated, where, for each scene configuration, the vehicles are moved to new positions. From the total amount of real data, the same five sequences, totaling 3,000 real point clouds, are used for the testing split for all tests performed. The five sequences are chosen such that the distribution of points per model is as even as possible across all possible tractor and combine models captured in the dataset. The remaining 6,000 real point clouds, which come from different vehicle configurations than those used in the testing set, are then used in the training of the real-only baseline models.

Each point cloud consists of around 50,000 points on average, where each point is labeled into separate classes. The dataset has three different classes: tractor, combine harvester, and other. Additionally, the average class-wise point distribution for each point cloud is 6.5\% tractor, 12.6\% combine harvester and 80.9\% other. To annotate the point cloud the static environment used to capture the data is leveraged to create a combined point cloud for each sequence using KISS-ICP \cite{vizzo2023ral}. Clustering is then applied to each vehicle in the combined points cloud and these clusters are used to annotate the individual point clouds in the sequence.

\subsection{Synthetic Data Generation} \label{subsec:synthetic_data_generation}
Multiple methods can be used to generate synthetic point cloud data, as mentioned in Section \ref{subsec:synth_data_related}. Xiao \etal \cite{Xiao2021SynLiDAR}, showed that synthetic data, modified to close the sim-to-real gap, outperformed the purely synthetic data. This motivates the use of a simulated environment where points can be sampled in a LiDAR pattern compared to uniformly sampling points from the surfaces of the meshes as done by Ma \etal \cite{ma2020semantic}. When producing synthetic data using our simulation, data is only generated for the three different classes also available in the real-world LiDAR captured dataset.

\textit{Mesh generation:} To simulate the environment, which synthetic point clouds will be extracted from, it is of utmost importance to obtain the best possible meshes of the vehicles in the scene. The better the meshes resemble the real-world vehicles, the better the simulated LiDAR will be at sampling synthetic point clouds close to an actual real-life scene. The method for generating the meshes starts with capturing images of a scene where the vehicle is focused in the middle. This was done by capturing a video with a drone flying slowly around the vehicle, then sampling images from the video at a consistent interval.

\begin{figure}
    \centering
    \includegraphics[width=0.7\linewidth]{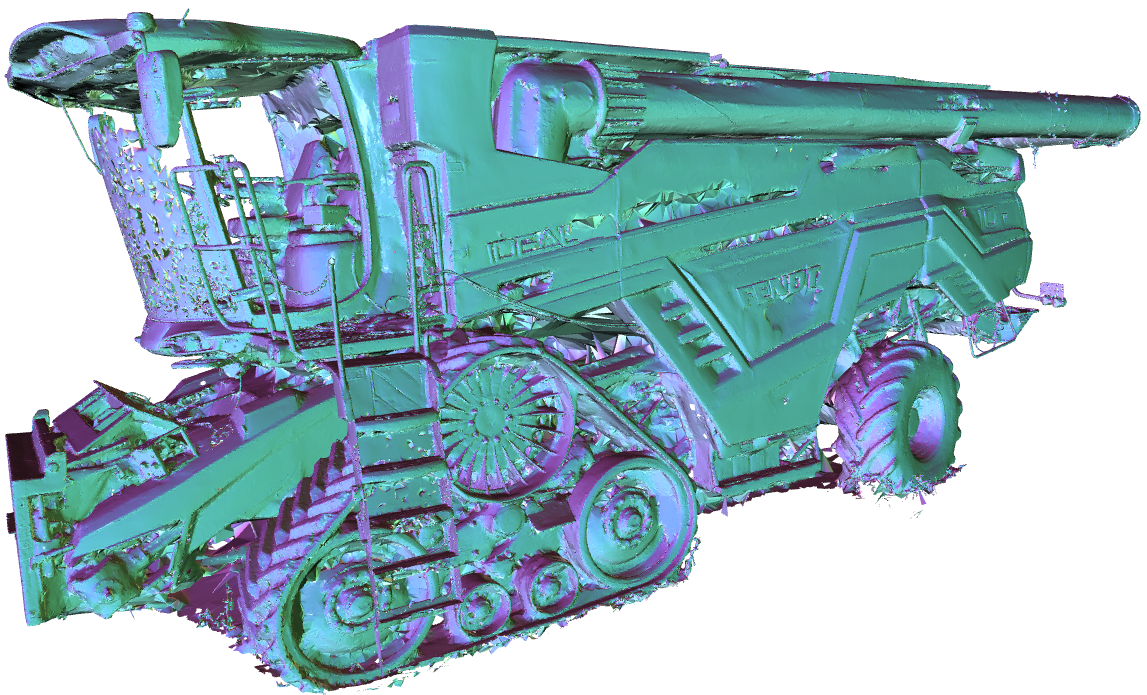}
    \caption{Figure shows a generated mesh where a single combine harvester has been cropped out, yielding a usable mesh for the Gazebo simulation.}
    \label{fig:combine_mesh}
\end{figure}

Using the captured images, the initial sparse point cloud is computed using the Structure-from-Motion (SfM) implementation in COLMAP \cite{schoenberger2016sfm}. Following this, the sparse SfM point cloud is utilized for the 3DGS mesh extraction algorithm. Through initial experimentation, it was found that, out of SuGaR \cite{guedon2024sugar}, 2DGS \cite{huang20242d} and GOF \cite{yu2024gaussian}, the meshes extracted using GOF yielded the best results, with the highest degree of fidelity. Meshes were then generated for all the different vehicles, which includes seven different tractor models for the tractor class of the dataset, and three different combine harvesters for the combine class. An example of a combine harvester mesh can be seen in Figure \ref{fig:combine_mesh}. A bit of post-processing is done on the meshes, since the output from the 3DGS mesh extraction is one big mesh of the whole scene. The post-processing comprises of cropping out everything except the specific vehicle in question. The meshes are then employed in the Gazebo simulation to generate the synthetic point clouds. Additionally, to achieve a simulation environment resembling the real world as best as possible, grass and other miscellaneous object meshes from the 3DGS meshes are utilized as well.

\textit{Simulation Environment:} Gazebo is an open-source simulation tool designed to simulate robotic applications. It is built around the Ogre2 engine, and has an integrated LiDAR plugin. 
Using the LiDAR plugin, it is possible to simulate any real LiDAR sensor geometrically. For the data generation the Ouster OS0-128 LiDAR is simulated, as it was used to capture the real dataset. Due to this, the synthetically extracted point clouds also end up being roughly the same size, with around 62,000 points on average per point cloud, with the average class-wise point distribution for each point cloud being 88.3\% other, 5.0\% tractor and 6.7\% combine harvester.

\begin{figure}
    \centering
    \includegraphics[width=\linewidth, trim={0 0cm 0 0cm}, clip]{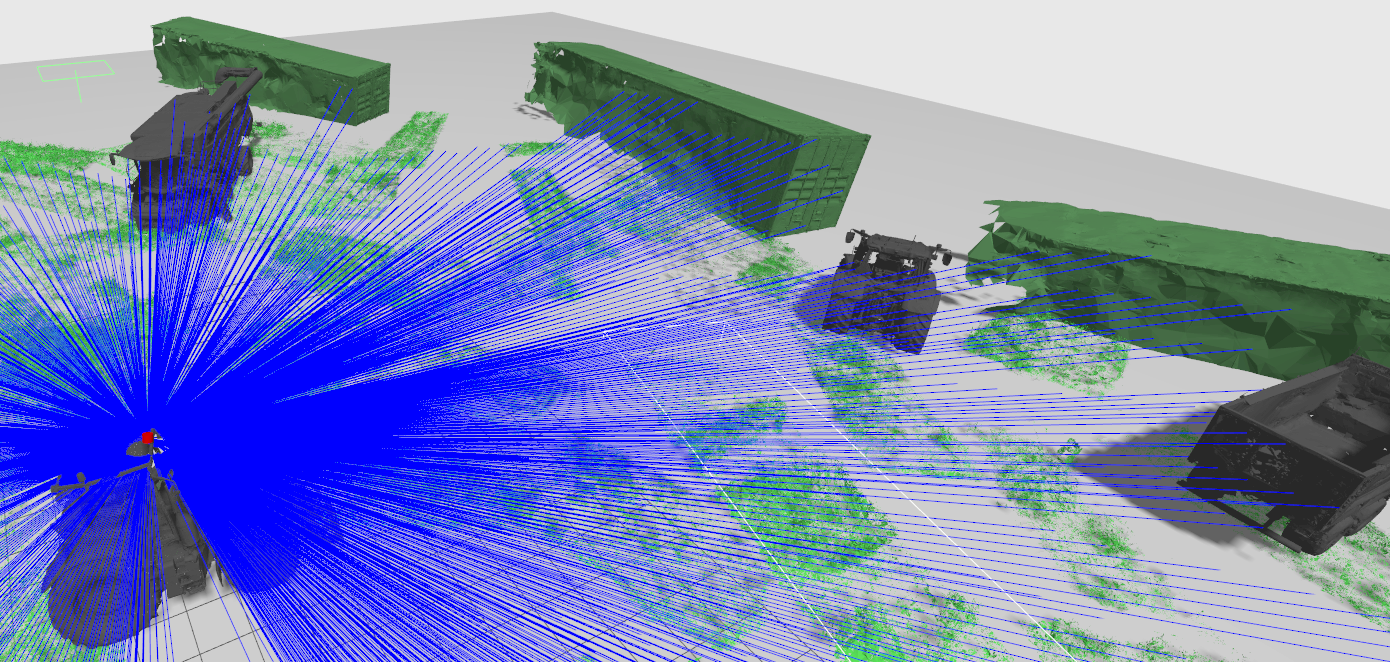}
    \caption{The Gazebo simulation where the target assets have moved to random positions. The LiDAR sensors position is marked by the red cylinder. The blue rays visualize a sparse version of the LiDAR rays}
    \label{fig:gazebo_random_movements}
\end{figure}

\begin{figure}
    \centering
    \includegraphics[width=\linewidth]{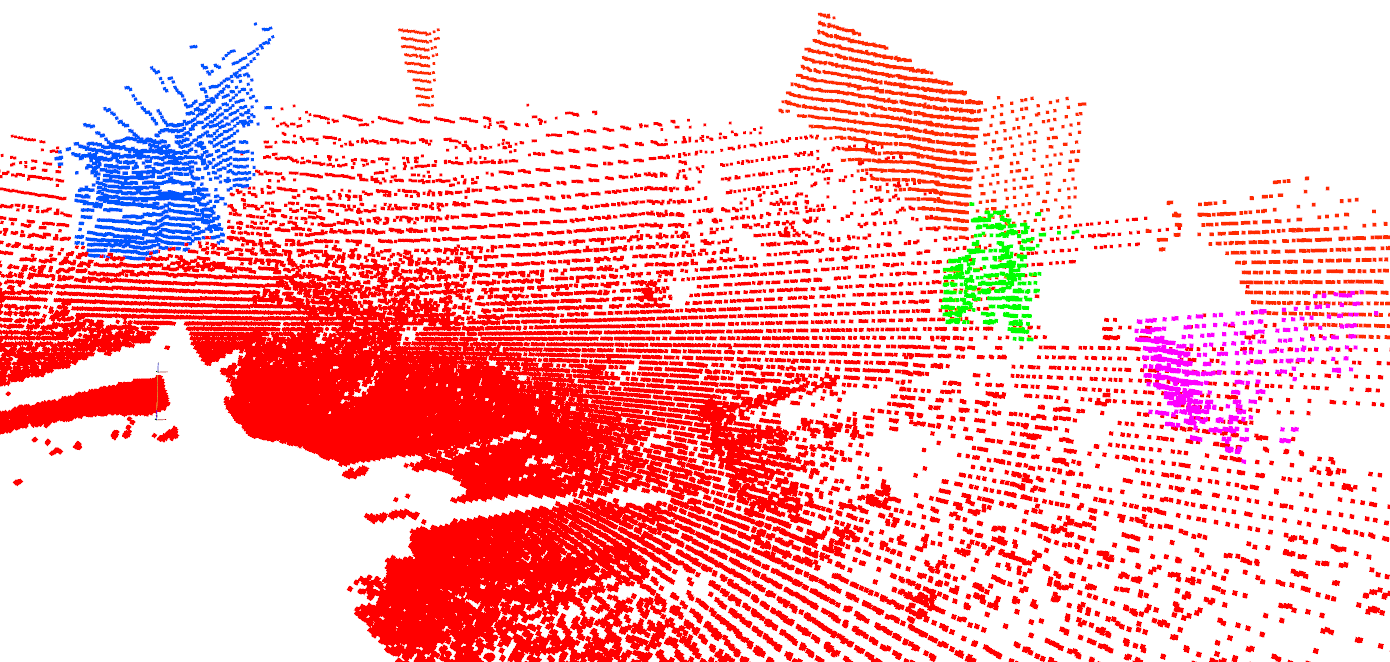}
    \caption{An annotated point cloud obtained from the gazebo simulation containing tractors (green), combine harvesters (blue), trailers(pink), and other (red).}
    \label{fig:rviz_sim_point_cloud}
\end{figure}

To use Gazebo as a dataset generator, a custom plugin is used to move the meshes of different classes and the LiDAR to random positions. To avoid meshes being placed out of range for the LiDAR or overlapping with one another, placement rules were made for the meshes and the LiDAR sensor. The result of this is depicted on Figure \ref{fig:gazebo_random_movements} and \ref{fig:rviz_sim_point_cloud}.

\section{Experimental Design}
To evaluate the effectiveness of our synthetic point cloud generation method, we conduct a series of experiments focused on 3D point cloud segmentation. This section focuses on outlining the 3D point cloud segmentation models used, the conducted tests and the dataset compositions used.

\subsection{3D Point Cloud Segmentation Model Selection}
All tests performed will be done on three different models, specifically, PointNet++, PTv3 and OACNN. These models were chosen on the basis that all three approach the point cloud segmentation problem quite differently in their point cloud processing and model architectures. The hyperparameters used for the three different models are based upon the original paper implementations of the respective models. We performed a small search on the learning rate and number of epochs, as shown in the supplementary materials.

\subsection{Synthetic Only Training} \label{subsec:synthetic_only_test}
Recent advances in Gaussian splatting scene representation have allowed for easy generation of highly detailed meshes of custom objects which are very difficult and time consuming to model from scratch. The mesh representation can be used to generate highly realistic semantic segmentation LiDAR datasets for training, as it is easy to model custom LiDAR scanning patterns. Given the relative ease in generating synthetic datasets, it is interesting to test the performance when only utilizing a synthetically generated dataset to train a segmentation model. \\
To assess the models' ability to generalize to real scenes, when trained exclusively on synthetic point clouds, a test is proposed. This test also assesses how accurate the simulation replicates real-world environments. The models are trained on 10,000 synthetic point clouds and validated on 2,000 synthetic only point clouds, to ensure the models trained using this approach, have never been exposed to any real point clouds until it is tested. Lastly, the models are tested on the test set containing 3,000 real point clouds, outlined in Section \ref{subsec:lidar_captured_dataset}.

\subsection{Tractor Generalization Test} \label{subsec:generalization_test}
It can be hypothesized that this method of using synthetically generated data can generate otherwise hard to acquire datasets of custom objects we have physical access to, with sufficient fidelity, such that the models trained on the synthetic datasets will be able to generalize to new unseen objects of the same semantic category, when captured with real LiDAR sensors.\\
To test this hypothesis, multiple synthetic datasets were created in which a tractor was removed from the available assets during the generation process. As a result, each dataset lacked one specific tractor model. 
Individual segmentation models were then trained on these modified datasets and had their performance evaluated on the same 3,000 real point clouds as previous tests, which contains all the different tractor models.

\subsection{Extended Synthetic Dataset Test} \label{subsec:extended_dataset_test}
Given the ease of generating synthetic point cloud datasets, it is valuable to investigate how expanding the synthetic dataset used in Section \ref{subsec:synthetic_only_test} affects the performance of models trained exclusively on synthetic data. To explore this, we conduct an experiment in which the number of synthetic point clouds used for training is significantly increased, from the initial 10,000 to approximately 65,000, while keeping the model architecture and training pipeline consistent.

\subsection{Prediction Visualization for Analysis}
As the testing scenes are static, the transforms between each point cloud can be found using KISS-ICP \cite{vizzo2023ral}. The transforms can be used to align the point clouds for a testing sequence creating a combined point cloud. Visualizing this point cloud provides insight into how the models generally segment the point clouds. This is done to qualitatively asses the segmentation quality of the different models. 
\section{Results} \label{sec:results}

\subsection{Synthetic Only Test} \label{subsec:synthetic_only_test_results}
Table \ref{tab:zeroshot_performance} presents the results for the synthetic only test, which tests the performance of training and validating the models without any real data. It can be seen that the mIoU across the models is on average $2.99$ percentage points worse for the synthetic only models compared to the real only. The class specific IoU results reveal that all models struggle the most with the "tractor" class, which is consistent with the class-wise point distributions described in Section \ref{subsec:synthetic_data_generation}.


\begin{table}
\centering
\resizebox{\columnwidth}{!}{%
\begin{tabular}{l|cc|cc|cc}
        & \multicolumn{2}{c|}{PointNet++} & \multicolumn{2}{c|}{PTv3} & \multicolumn{2}{c}{OACNN} \\
Class   & Baseline       & Synth only     & Baseline    & Synth only  & Baseline    & Synth only  \\ \hline
Tractor & 0.5580         & 0.6430         & 0.8675      & 0.7957      & 0.9052      & 0.7755      \\
Combine & 0.8180         & 0.7580         & 0.9145      & 0.8853      & 0.9273      & 0.8857      \\
Other   & 0.9720         & 0.9670         & 0.9878      & 0.9790      & 0.9885      & 0.9793      \\ \hline
mIoU    & 0.7824         & 0.7893         & 0.9232      & 0.8867      & 0.9403      & 0.8802      \\ \hline
\end{tabular}%
}
\caption{Table displays the individual IoU's and mIoU for each model traing two different datasets. Baseline is trained on all the available real data, and Synth only, is a synthetic only dataset with 10k point clouds.}
\label{tab:zeroshot_performance}
\end{table}

\subsection{Tractor Generalization Test} \label{subsec:generalization_test_results}
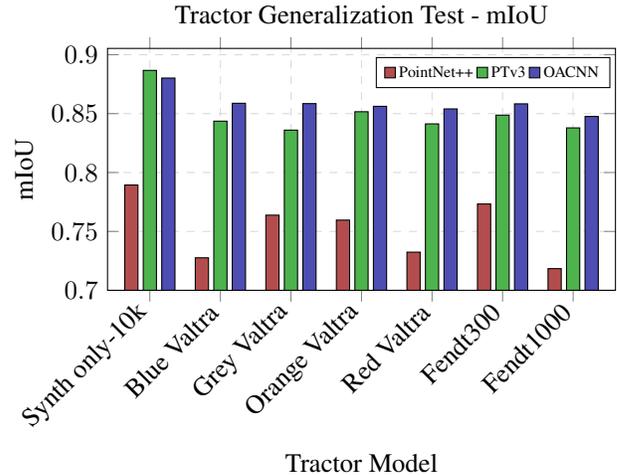
\begin{figure}
  \begin{center}
    \begin{tikzpicture}
      \begin{axis}[
          width=\columnwidth,
          height=4.8cm,
          ybar,
          bar width=5pt,  
          grid=major,
          grid style={dashed,gray!30},
          xlabel=Tractor Model,
          ylabel=mIoU,
          symbolic x coords={Synth only-10k, Blue Valtra, Grey Valtra, Orange Valtra, Red Valtra, Fendt300, Fendt1000},
          xtick=data,
          ymin=0.70, 
          enlarge x limits=0.1,  
          xticklabel style={rotate=45, anchor=east},
          legend style={at={(0.99,0.90)}, anchor=east, nodes={scale=0.6}, legend columns=3},
          legend image code/.code={
            \draw[#1,fill=#1] (-0.02cm,-0.07cm) rectangle (0.12cm,0.07cm);
          },
          title=Tractor Generalization Test - mIoU,
        ]


        \addplot[fill=red!40!gray] coordinates { 
            (Synth only-10k, 0.7894)
            (Blue Valtra, 0.7276) 
            (Grey Valtra, 0.7638) 
            (Orange Valtra, 0.7596)
            (Red Valtra, 0.7324)
            (Fendt300, 0.7733) 
            (Fendt1000, 0.7184)
        };
        \addlegendentry{PointNet++}
        
        \addplot[fill=green!40!gray] coordinates { 
            (Synth only-10k, 0.8867) 
            
            (Blue Valtra, 0.8436)
            
            (Grey Valtra, 0.8360)
            
            (Orange Valtra, 0.8516)
            
            (Red Valtra, 0.8413)
            
            (Fendt300, 0.8487) 
            
            (Fendt1000, 0.8379)
        };
        \addlegendentry{PTv3}

        \addplot[fill=blue!40!gray] coordinates { 
            (Synth only-10k, 0.8802) 
            
            (Blue Valtra, 0.8588)
            
            (Grey Valtra, 0.8585)
            
            (Orange Valtra, 0.8562)
            
            (Red Valtra, 0.8540)
            
            (Fendt300, 0.8583) 
            
            (Fendt1000, 0.8476)
        };
        \addlegendentry{OACNN}

      \end{axis}
    \end{tikzpicture}
    \caption{Figure shows the mIoU results for each test in which the specified tractor model was not included in the training/validation data. The "Synth only-10k" entry contains all the tractor models.}
    \label{fig:tractor_gen}
  \end{center}
\end{figure}

The results from testing the ability to generalize across individual tractor models within the same semantic class are presented in Figure \ref{fig:tractor_gen}. The figure displays each individual model that has been trained on a dataset missing the displayed tractors, these models are also compared to the synthetic only model, seen in Section \ref{subsec:synthetic_only_test_results}, as a baseline comparison which has been trained on all available tractors. The results show a general tendency towards a drop in performance when missing a tractor by, on average, $3.65\%$ percentage points.

The mean of all the unseen tractor IoU's is used as a metric to compare the mean tractor IoU for the seen tractors, as this gives an image of how well the models generalize to unseen tractors compared to seen tractors. This is represented in Table \ref{tab:tractor_gen_iou}, where it can be seen that there is a average drop in accuracy for unseen tractors, but the effect is limited, which shows that the models are able to generalize on the tractor class. 

\begin{figure}
    \centering
    \includegraphics[width=\linewidth]{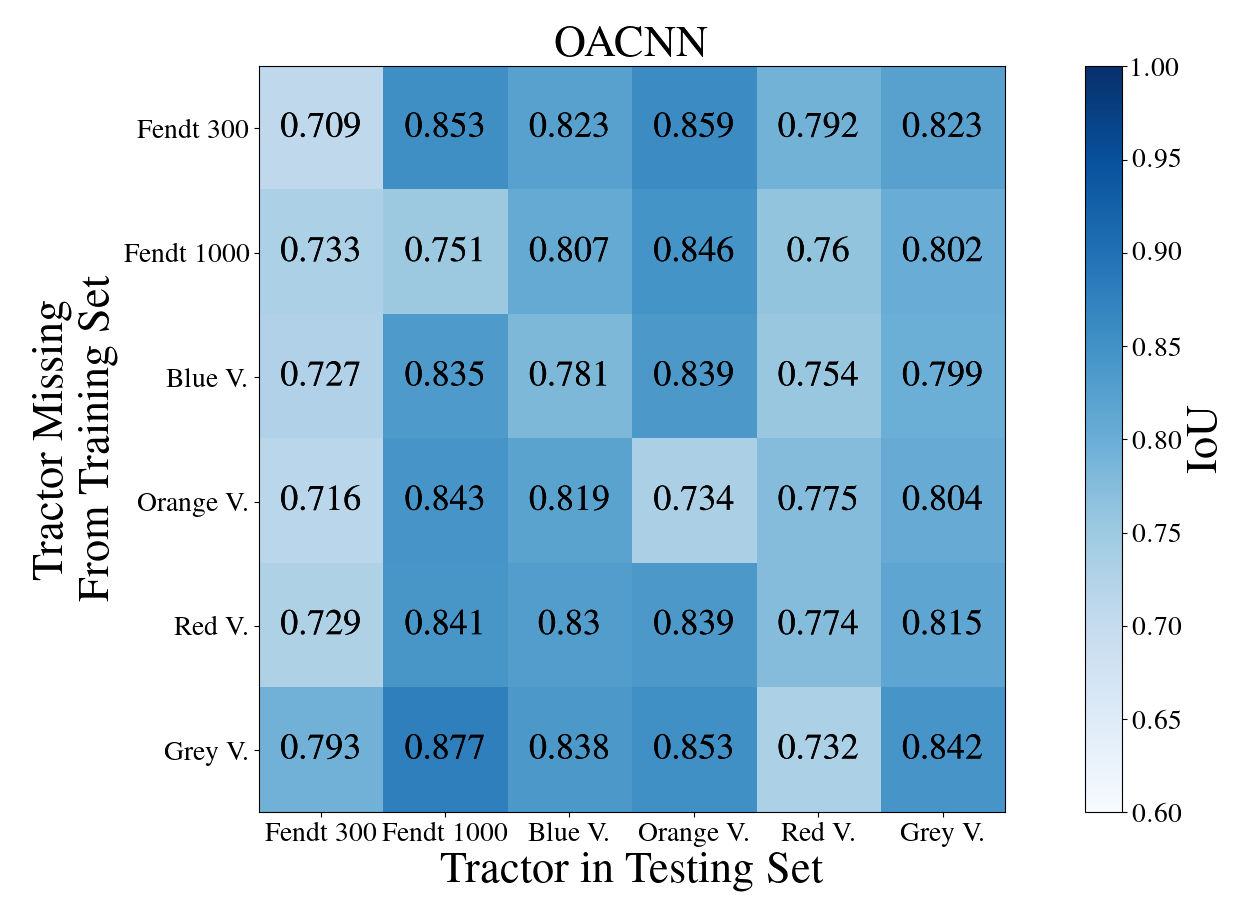}
    \caption{OACNN IoU of tractors when not in training set}
    \label{fig:OACNN_tractor_gen_acc_v2}
\end{figure}

An overall view of the difference between the model having seen the tractor in the training set is presented in Figure \ref{fig:OACNN_tractor_gen_acc_v2}. The matrix displays a small correlation between the tractor missing in the training set, and a lower IoU of the tractor in the testing set, however it is not always the case.

\begin{table}
    \centering
    \resizebox{0.9\columnwidth}{!}{%
    \begin{tabular}{l|c|c}
        Model & PTv3 & OACNN\\ \hline
        Mean IoU of unseen tractors &   0.7273 &  0.7652\\
        Mean IoU of seen tractors&  0.7603 &   0.8052\\ \hline
    \end{tabular}
    }
    \caption{Figure shows the mean IoU of tractor models not included in the training process compared to tractors which are included in the training process.}
    \label{tab:tractor_gen_iou}
\end{table}

\subsection{Extended Synthetic Dataset Test} \label{subsec:extended_dataset_test_results}
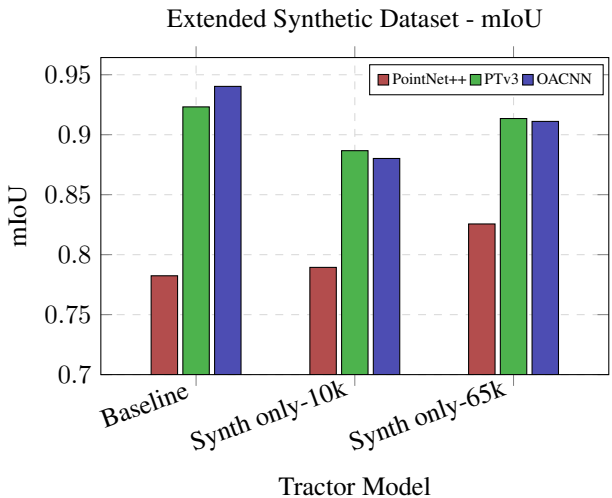
\begin{figure}[t]
  \begin{center}
    \begin{tikzpicture}
      \begin{axis}[
          width=\columnwidth,
          height=5.8cm,
          ybar,
          bar width=10pt,  
          grid=major,
          grid style={dashed,gray!30},
          xlabel=Tractor Model,
          ylabel=mIoU,
          ytick={0.70, 0.75, 0.80, 0.85, 0.90, 0.95},
          symbolic x coords={Baseline, Synth only-10k, Synth only-65k},
          xtick=data,
          ymin=0.70,
          enlarge x limits=0.3,  
          xticklabel style={rotate=20, anchor=east},
          legend style={at={(0.99,0.93)}, anchor=east, nodes={scale=0.6}, legend columns=3},
          legend image code/.code={
            \draw[#1,fill=#1] (-0.02cm,-0.07cm) rectangle (0.12cm,0.07cm);
          },
          title=Extended Synthetic Dataset - mIoU,
        ]


        \addplot[fill=red!40!gray] coordinates {
            (Baseline, 0.7824)
            (Synth only-10k, 0.7894)
            (Synth only-65k, 0.8256)
        };
        \addlegendentry{PointNet++}
        
        \addplot[fill=green!40!gray] coordinates { 
            (Baseline, 0.9232)
            (Synth only-10k, 0.8867)
            (Synth only-65k, 0.9135)
        };
        \addlegendentry{PTv3}

        \addplot[fill=blue!40!gray] coordinates { 
            (Baseline, 0.9403)
            (Synth only-10k, 0.8802)
            (Synth only-65k, 0.9111)
        };
        \addlegendentry{OACNN}

      \end{axis}
    \end{tikzpicture}
    \caption{Figure shows the comparison between real only trained baseline models and synth only 10k models, from Table \ref{tab:zeroshot_performance}, with synth only trained models using 65k synthetic point clouds.}
    \label{fig:extended_dataset}
  \end{center}
\end{figure}

The results of the test can be seen on Figure \ref{fig:extended_dataset}, which displays the results marked with the "Synth only-65k" label. It can be seen that performance increases significantly with a larger dataset, where PTv3 and OACNN achieve over 90\% in mIoU when trained only on 65k synthetic point clouds. Additionally, it can be seen that both of the two synthetically trained PointNet++ models outperform the baseline, which is trained only on real data. This could potentially be due to the lack of augmentations in the PointNet++ implementation as opposed to PTv3 and OACNN which use several augmentation techniques.

\subsection{Qualitative Analysis} 
\begin{figure*}
    \begin{subfigure}[t]{\linewidth}
    \centering
    \includegraphics[width=0.8\linewidth]{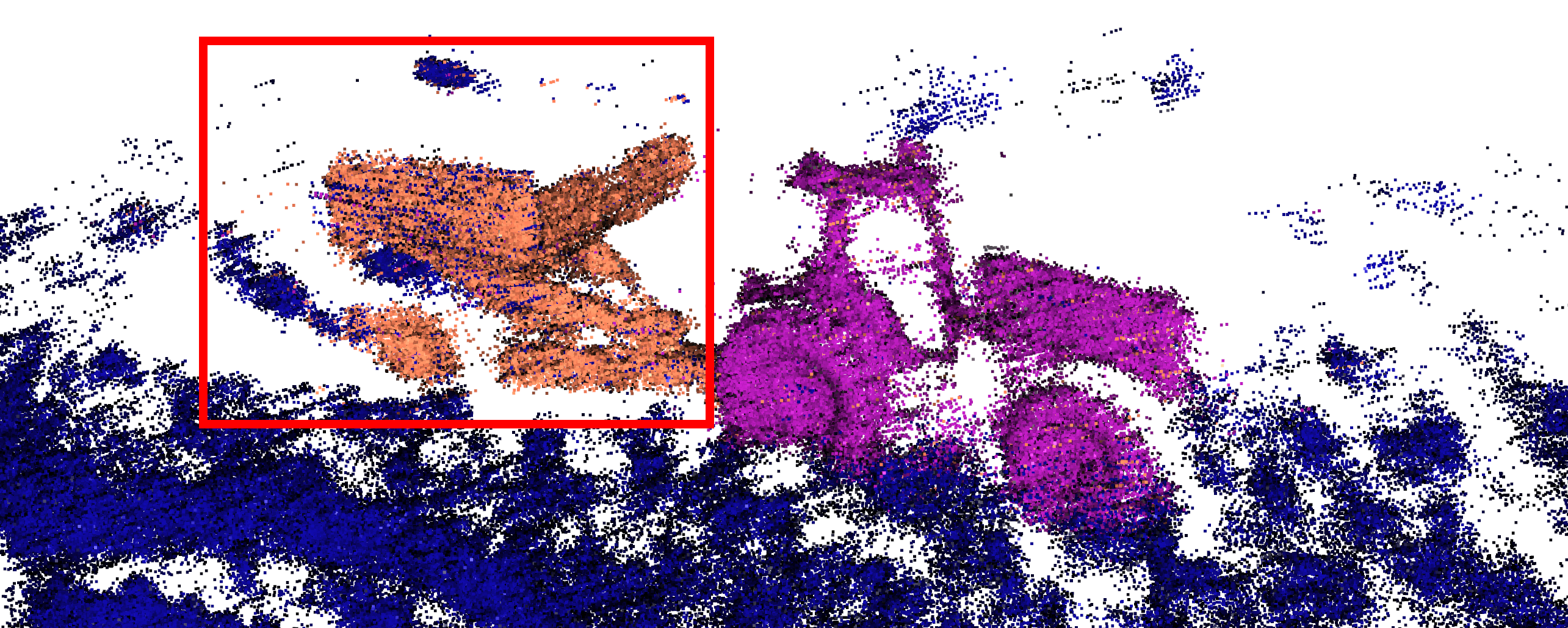}
    
    \caption{\textbf{OACNN Baseline model point predictions}. Notice how the trailer is incorrectly predicted to be a combine harvester, as highlighted by the red \legendsquare{redbbx} bounding box.}
    \label{fig:oacnn_basline_tractor_and_trailer}
    \end{subfigure}
    \begin{subfigure}[t]{\linewidth}
    \centering
    \includegraphics[width=0.8\linewidth]{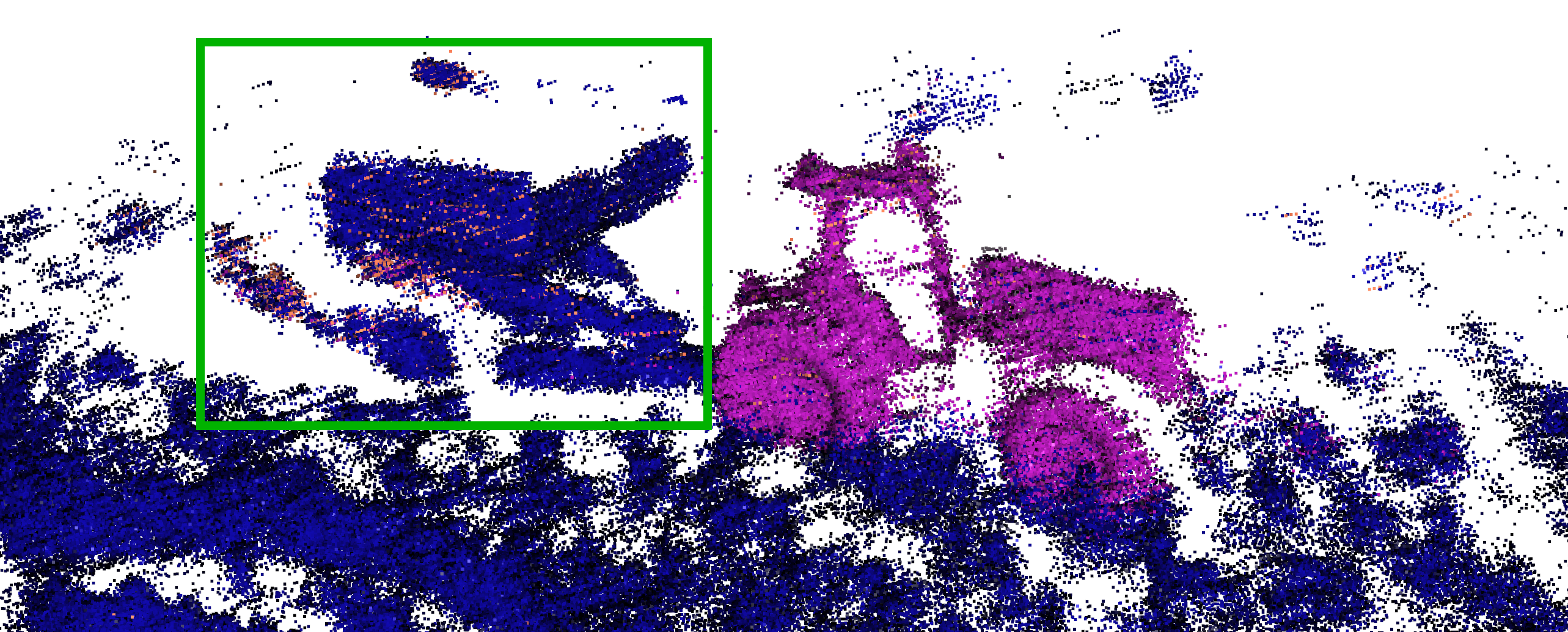}
    
    \caption{\textbf{OACNN Synth only-65k model point predictions}. Notice how the trailer is correctly predicted to be of the ''other``, as highlighted by the green \legendsquare{greenbbx} bounding box.}
    \label{fig:oacnn_65k_tractor_and_trailer}
    \end{subfigure}
    \caption{We compare the point predictions of a OACNN network trained in the baseline configuration versus trained with the Synth only-65K configuration. The scene depicts a tractor and a trailer. The point colors indicate the predicted class: \legendsquare{ground} other, \legendsquare{tractor} tractor, and \legendsquare{combine} combine harvester}
\end{figure*}

Using the models trained in Section \ref{subsec:synthetic_only_test} and \ref{subsec:extended_dataset_test}, all seen in Figure \ref{fig:extended_dataset}, combined point clouds can be used to find the places where they differ, that are not apparent from IoU and mIoU numbers. The first example of where they differ is in models under represented in the real training data. This is seen on Figure \ref{fig:oacnn_basline_tractor_and_trailer}, where the trailer is completely misclassified as a combine harvester. On Figure \ref{fig:oacnn_65k_tractor_and_trailer} the model trained on the largest synthetic dataset handles the trailer perfectly and classifies it as the correct "other" class in the majority of the points. However, we find that the the synthetic models fail segmenting when tall grass is present, as shown in the supplementary materials.

\section{Discussion}

Throughout the experiments in this paper, we have not focused on the performance of the individual segmentation models, but rather on determining the influence of synthetically generated data on the performance of point cloud segmentation models. As a result, the performance in our tests could potentially be improved by modifying and fine-tuning the hyperparameters. One of the tests used to gauge the influence of synthetic data on the segmentation models, was the synthetic only test. This test showed that training only on synthetic data, could potentially be a feasible solution in domains where data is hard to acquire. Especially with larger synthetic datasets, as seen in Section \ref{subsec:extended_dataset_test_results}, where OACNN and PTv3, trained on 65k point clouds, compared similarly to the baseline models which were trained on real data. The qualitative analysis revealed that a trailer in the testing dataset was misclassified as a combine with the real-only OACNN baseline. This was not the case with the OACNN model trained on 65k synthetic point clouds. The big difference between the training datasets of these two models, apart from the size, is the distribution of points per mesh/model. In the synthetically generated dataset, the trailer is much more common than in the real dataset. This, along with the enlarged training dataset, could be the reason for the increased performance. Additionally, the qualitative analysis also revealed that the real data helps with the classification of the tall grass in the background of the point clouds, as shown in the supplementary materials. This is most likely due to the similarity of the testing- and training dataset as they were captured on the same field, which would explain why the baseline outperformed the synthetic-only model in the area with tall grass.
\section{Conclusion} \label{sec:conclusion}
In this paper we have proposed and evaluated a novel pipeline to efficiently train point cloud segmentation models in scenarios with limited real data. The pipeline leverages GOF, a state of the art technique for mesh extraction, to obtain meshes that accurately represent target vehicles in high detail.

A simulation environment, utilizing the high quality meshes, was developed to efficiently generate LiDAR datasets suitable for semantic segmentation. Secondly, a semi-automatic annotation technique was developed, to annotate the data from the real LiDAR. 

Three models were tested, namely: Point Transformer V3 \cite{wu2024pointtransformerv3simpler}, Omni-Adaptive Sparse CNN \cite{peng2024oacnnsomniadaptivesparsecnns}, and Pointnet++ \cite{qi2017pointnetdeephierarchicalfeature}. 
Multiple tests were carried out for the different models exploring different ways that synthetic data could be used to train point cloud segmentation models. The synthetic only test, seen in Section \ref{subsec:synthetic_only_test_results}, shows the potential of training models purely on synthetic data, with OACNN and PTv3 achieving $+88\%$ mIoU. Additionally, the extended dataset test, seen in Section \ref{subsec:extended_dataset_test_results}, showed that expanding the dataset significantly improved the mIoU for all the models, with OACNN and PTv3 now surpassing $+91\%$ in mIoU, almost comparable to the baseline trained on real data.
Secondly it was shown that the model is able to generalize well to unseen tractor models when trained only on synthetic data, with a mean performance drop in IoU for unseen tractors of $3.65$ percentage points from Table \ref{tab:tractor_gen_iou}. 
Thirdly, qualitative analysis showed that in some cases the models trained on synthetic data had more desirable predictions, which is also a strong argument for synthetic data, as this is presumably caused by the perfect annotations that are acquired when using the proposed pipeline. 
Finally, with the results gathered from all the experiments, it is shown that using the proposed novel pipeline for synthetic data acquisition, is a viable solution when gathering data for training point cloud segmentation models in uncommon domains. While this paper focused on a single domain, there is strong evidence which suggests that other domains would benefit from using a similar data generation pipeline.
Moving forward, it would be interesting to see the effects of this method on more common domains, such as semanticKITTI \cite{behley2019semantickittidatasetsemanticscene}, to evaluate its effect on well-known datasets and gain insights into the drawbacks and benefits.

\textbf{Funding} This research was funded by Innovation Fund Denmark, grant number 3129-00060B.



{
    \small
    \bibliographystyle{ieeenat_fullname}
    \bibliography{mybib}
}
\newpage
\appendix
\clearpage
\setcounter{page}{1}
\maketitlesupplementary

\section{Model and Training Tuning}
Given the objective of the paper has not been to fine-tune the individual models to obtain the best possible performance, model training efficiency was valued highly when weighing performance against training time for the hyperparameter selection, and for the most part, hyperparameters have been chosen based solely on the model's original paper implementations \cite{qi2017pointnetdeephierarchicalfeature, wu2024pointtransformerv3simpler, peng2024oacnnsomniadaptivesparsecnns}. Thus, as a last test it also seemed interesting to observe the performance differences when modifying the base hyperparameters. The tests focused on three key hyperparameters: the number of training epochs, the learning rate and the degree of point cloud downsampling.

\subsection{Epoch Tuning} \label{sec:epoch-tuning}
The default number of epoch used throughout the test has been 20, thus when varying the number of epochs it was chosen to test \{10, 20, 30\} epochs. The results can be found in Table \ref{tab:epoch_ablation}. As the results suggest, increasing the number of training epochs helps performance quite notably, and from the training process it seems like the models haven't quite converged, and thus could improve even further given additional epochs.

\begin{table}
\centering
\resizebox{0.8\columnwidth}{!}{%
\begin{tabular}{c|c|c|c}
Epochs & PointNet++ & PTv3 & OACNN \\ \hline
10     & 0.7608    & 0.8549    & 0.8708     \\ \hline
20     & 0.7844    & 0.8867    & 0.8802     \\ \hline
30     & 0.7894    & 0.8911    & 0.8917     \\ \hline
\end{tabular}%
}
\caption{Table displays the mIoU of the three different models when trained for three different epoch amounts.}
\label{tab:epoch_ablation}
\end{table}

\subsection{Learning Rate Tuning} \label{sec:lr-tuning}
The learning rate is usually the most impactful hyperparameter when changed, in this case a sweep was conducted where the base learning rate used for all other model's tests was scaled by \{0.1, 1.0, 10.0\}. The results can be seen in Table \ref{tab:lr_ablation}. Interestingly it seems the PTv3 model would benefit from a lowered learning rate, and when scaling it by 10, it would outright crash because of exploding gradients, further suggesting that the PTv3 model should have its learning rate lowered. The results for the OACNN model encourages the opposite, that it should be trained with a higher learning rate.

\begin{table}
\centering
\resizebox{0.8\columnwidth}{!}{%
\begin{tabular}{c|c|c|c}
LR scaling & PointNet++ & PTv3 & OACNN \\ \hline
0.1        & 0.6876     & 0.8936    & 0.8914     \\ \hline
1.0        & 0.7894     & 0.8867    & 0.8802     \\ \hline
10.0       & 0.6932     & N/A       & 0.8968     \\ \hline
\end{tabular}%
}
\caption{Table displays the mIoU of the three different models when trained on three different learning rate scales. The original implementation learning rates were 2e-3 with batch size 16 for both PointNet++ \cite{qi2017pointnetdeephierarchicalfeature} and OACNN \cite{peng2024oacnnsomniadaptivesparsecnns} and 5e-3 for PTv3 with batch size 12 \cite{wu2024pointtransformerv3simpler}. The conducted tests were run with batch size 32 for all models, consequently the learning rates were scaled proportially to this as well.}
\label{tab:lr_ablation}
\end{table}

\subsection{Point Cloud Downsampling} \label{sec:downsampling}
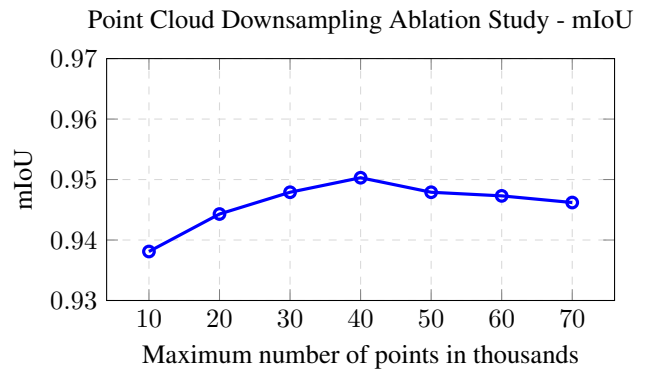
\begin{figure}
  \begin{center}
    \begin{tikzpicture}
      \begin{axis}[
          width=\columnwidth,
          height=4.8cm,
          grid=major,
          grid style={dashed,gray!30},
          xlabel=Maximum number of points in thousands,
          ylabel=mIoU,
          ymin=0.93, ymax=0.97,
          legend style={at={(0.98,0.18)},anchor=east,nodes={scale=0.7},},
          title=Point Cloud Downsampling Ablation Study - mIoU,
        ]
        
        \addplot[blue, mark=o, line width=1.2pt]
        coordinates {
            (10, 0.9381) (20, 0.9443) (30, 0.9479) (40, 0.9503) (50, 0.9479) (60, 0.9473) (70, 0.9462)
        };

      \end{axis}
    \end{tikzpicture}
    \caption{Figure shows the mIoU results for the OACNN model, trained on a real only dataset, with varying amounts of downsampling performed on the input point clouds.}
    \label{fig:downsampling_ablation}
  \end{center}
\end{figure}

To assess the trade-off between computational efficiency and segmentation performance, different point cloud densities were tested. A downsampled point cloud is obtained by performing a uniform random sampling without replacement from the original point cloud. The results can be seen in Figure \ref{fig:downsampling_ablation}. Mostly due to VRAM usage constaints, point clouds have been downsampled to 30.000 points for PointNet++ and 40.000 points for OACNN and PTv3 for all tests, and as the results display, this downsampling does not hinder the model from sufficiently learning the classwise point cloud representations, unless the point clouds are significantly downsampled.

\subsection{Mixed Training Dataset}\label{subsec:mixed_training}
To establish the effects of synthetic data on the 3D semantic segmentation models, we trained the three different models on a combination of synthetic data and real data, and then compared this to models trained only on real data. This was inspired by Ma et al. \cite{ma2020semantic}, Wang et al. \cite{wang2019automatic}, and Yue et al. \cite{yue2018lidar}. In total, the training dataset consists of 10,000 point clouds, with a 50/50 split of synthetic and real, as done by Yue et al. \cite{yue2018lidar}. The synthetic part is composed solely from unique point clouds, while the real part is composed of 20\%, 40\%, 60\%, 80\% and 100\% of all available unique real point clouds. The real point clouds are over-sampled to match the amount of synthetic point clouds, to avoid biasing the model towards the synthetic data. In total, five different datasets are produced, which individual models are trained upon, each of these datasets also contains the same validation split consisting of 1,200 real point clouds. For evaluation, all models are tested on the test set containing 3,000 real point clouds, outlined in Section \ref{subsec:lidar_captured_dataset}.

\begin{figure}
  \begin{center}
    \begin{tikzpicture}
      \begin{axis}[
          width=\columnwidth,
          height=4.8cm,
          grid=major,
          grid style={dashed,gray!30},
          xlabel=Amount of utilized real data,
          symbolic x coords={20\%, 40\%, 60\%, 80\%, 100\%},
          ylabel=mIoU,
          legend style={at={(0.98,0.18)},anchor=east,nodes={scale=0.7},},
          title=Mixed Training Dataset - OACNN - mIoU,
        ]
        
        \addplot[red, mark=o, dashed, line width=1.2pt]
        coordinates {
            (20\%, 0.8465) (40\%, 0.9191) (60\%, 0.9349) (80\%, 0.9378) (100\%, 0.9403)
        };
        \addlegendentry{Real}
        
        \addplot[blue, mark=o, line width=1.2pt]
        coordinates {
            (20\%, 0.9247) (40\%, 0.9451) (60\%, 0.9474) (80\%, 0.9498) (100\%, 0.9517)
        };
        \addlegendentry{Mixed}

      \end{axis}
    \end{tikzpicture}
    \caption{Figure shows the mIoU results for the OACNN model, trained on both a mixed dataset and a real only dataset, consisting of different amounts of different real point clouds.}
    \label{fig:mixed_results_miou_oacnn}
  \end{center}
\end{figure}
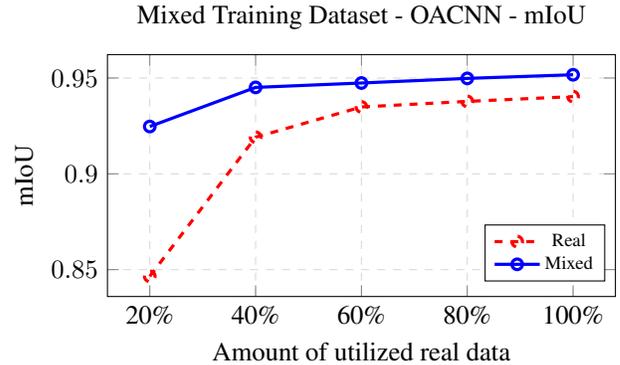

The test results evaluating the impact of the mixed dataset training compared to real-only dataset training are presented in Table \ref{tab:mixed_performance_increase} and Figure \ref{fig:mixed_results_miou_oacnn}. Table \ref{tab:mixed_performance_increase} compares the performance of the models, trained only on real data, to the models which were trained on a combination of real data and synthetic data. Additionally, the mean increase is shown, and it can be seen that the synthetic data improves performance, especially when only a small amount of real data is available. Figure \ref{fig:mixed_results_miou_oacnn} visually illustrates the mIoU performance of the OACNN model, comparing mixed dataset training with real-only dataset training as the number of different real point clouds increases.

\begin{table}
\centering
\resizebox{\columnwidth}{!}{%
\begin{tabular}{c|cc|cc|cc|c}
Sample &
  \multicolumn{2}{c|}{PointNet++} &
  \multicolumn{2}{c|}{PTv3} &
  \multicolumn{2}{c|}{OACNN} &
  Mean \\
percentage &
  \multicolumn{1}{l}{Real only} &
  \multicolumn{1}{l|}{Mixed} &
  \multicolumn{1}{l}{Real only} &
  \multicolumn{1}{l|}{Mixed} &
  \multicolumn{1}{l}{Real only} &
  \multicolumn{1}{l|}{Mixed} &
  increase \\ \hline
20\%  & 0.6982 & 0.7461 & 0.8369 & 0.8762 & 0.8465 & 0.9247 & 0.0551 \\ 
40\%  & 0.7215 & 0.7942 & 0.8983 & 0.9116 & 0.9191 & 0.9451 & 0.0373 \\ 
60\%  & 0.7406 & 0.7992 & 0.9096 & 0.9301 & 0.9349 & 0.9474 & 0.0305 \\ 
80\%  & 0.7505 & 0.7673 & 0.9065 & 0.9297 & 0.9378 & 0.9498 & 0.0173 \\ 
100\% & 0.7824 & 0.7894 & 0.9232 & 0.9328 & 0.9403 & 0.9517 & 0.0093 \\ \hline 
\end{tabular}%
}
\caption{Table displays the mIoU performance comparison between models trained on real dataset comprised of different amounts of real point clouds, and mixed datasets. The real only column shows baseline models trained on varying amounts of real data. Mean increase displays the average increase seen when using a mixed dataset compared to using only real. The data is presented for each sample percentage and each individual model.}
\label{tab:mixed_performance_increase}
\end{table}

\section{Combined point clouds with predictions}
\begin{figure*}[t]
    \centering
    \begin{tabular}{ccc}
        \begin{subfigure}{0.3\textwidth}
            \centering
            \includegraphics[width=\textwidth]{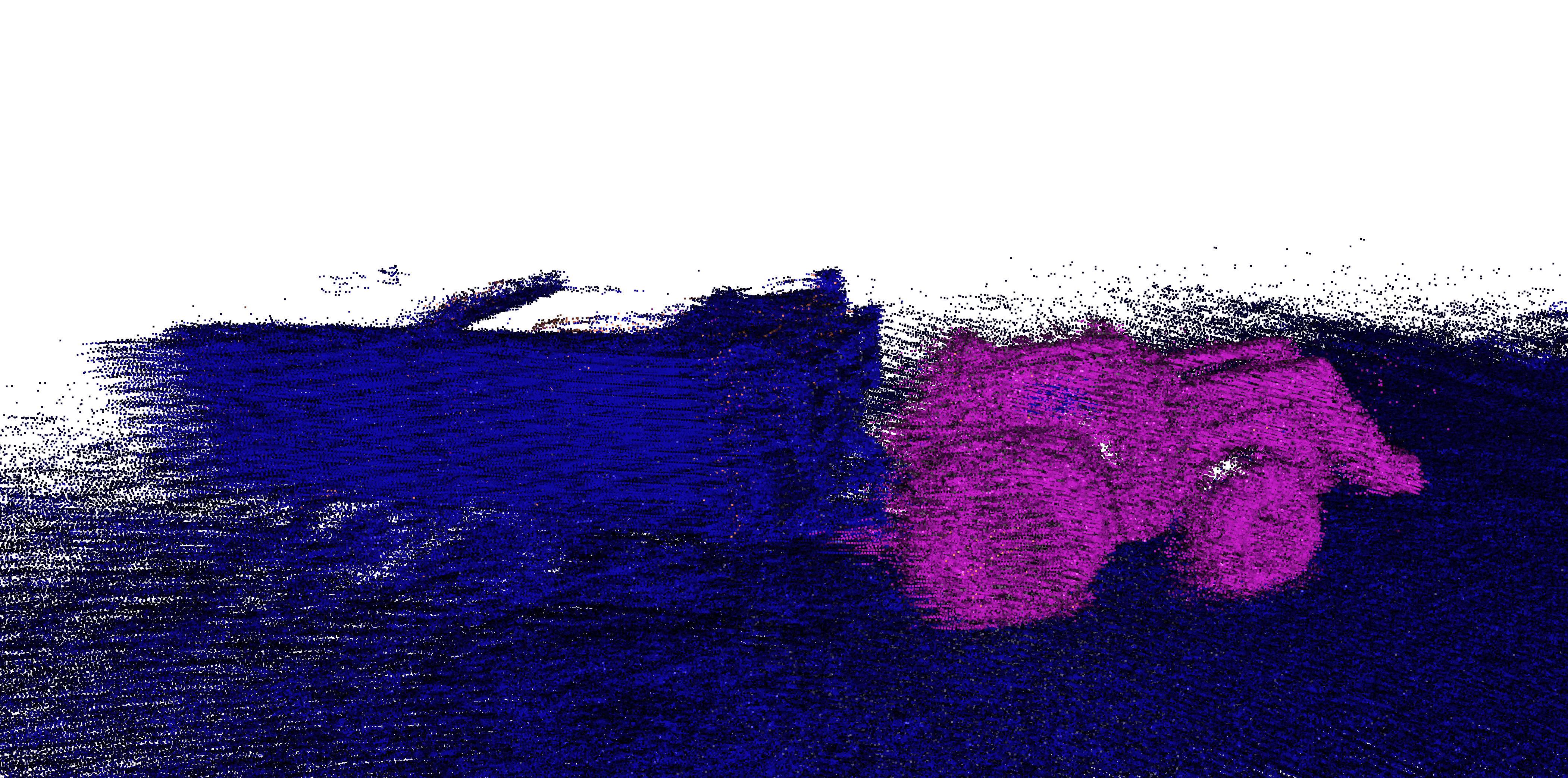}
        \end{subfigure} &
        \begin{subfigure}{0.3\textwidth}
            \centering
            \includegraphics[width=\textwidth]{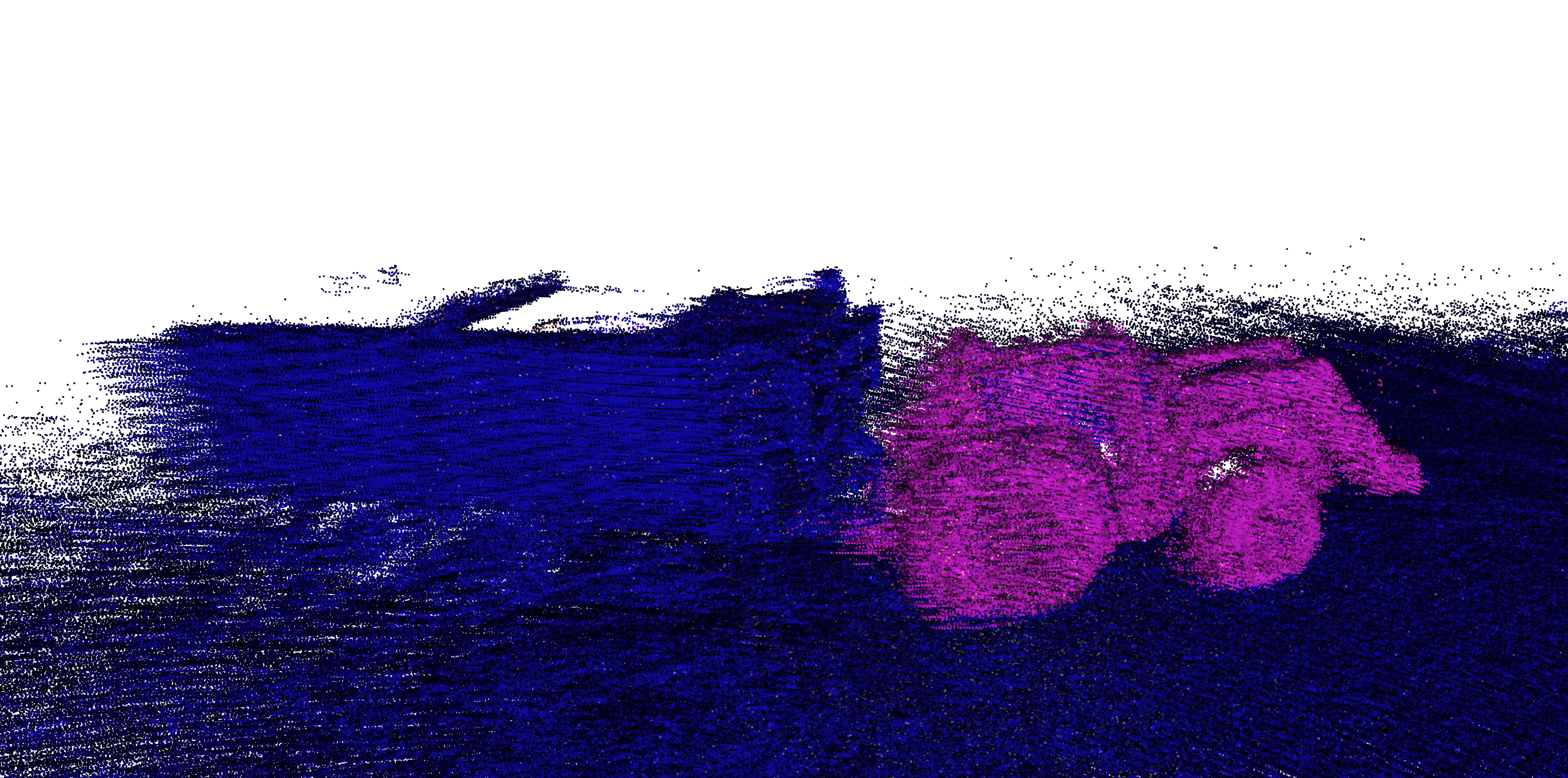}
        \end{subfigure} &
        \begin{subfigure}{0.3\textwidth}
            \centering
            \includegraphics[width=\textwidth]{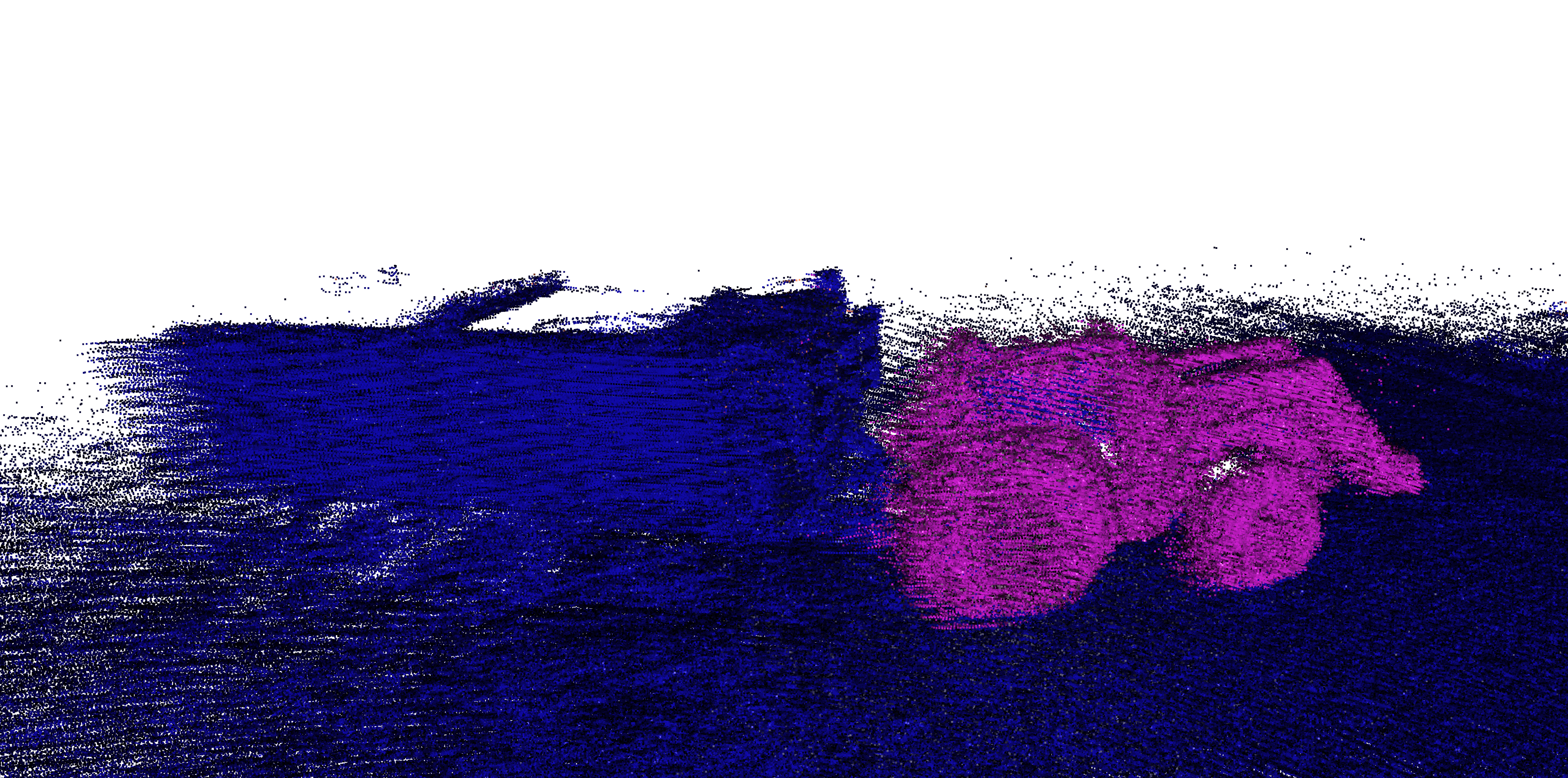}
        \end{subfigure} \\
        
        \begin{subfigure}{0.3\textwidth}
            \centering
            \includegraphics[width=\textwidth]{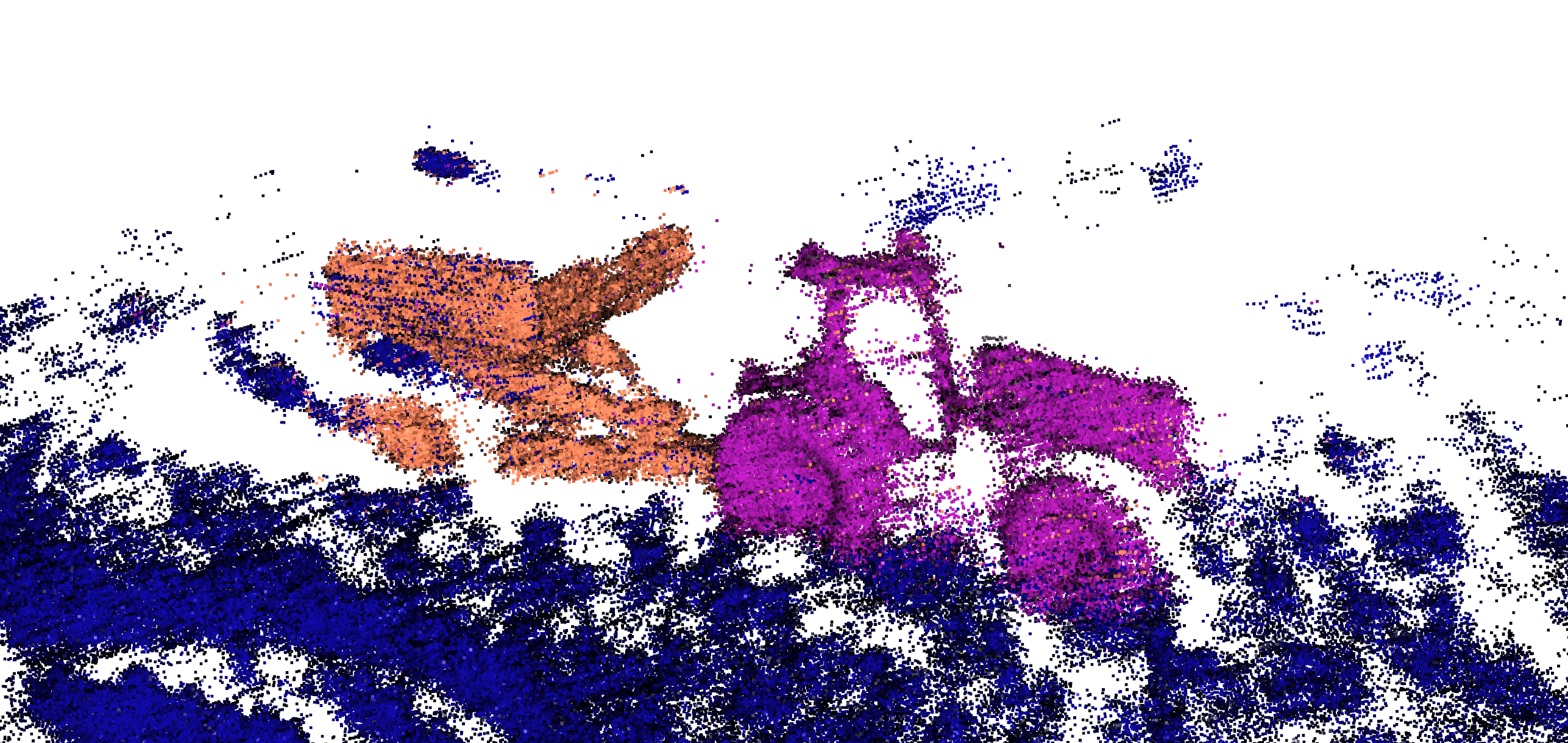}
        \end{subfigure} &
        \begin{subfigure}{0.3\textwidth}
            \centering
            \includegraphics[width=\textwidth]{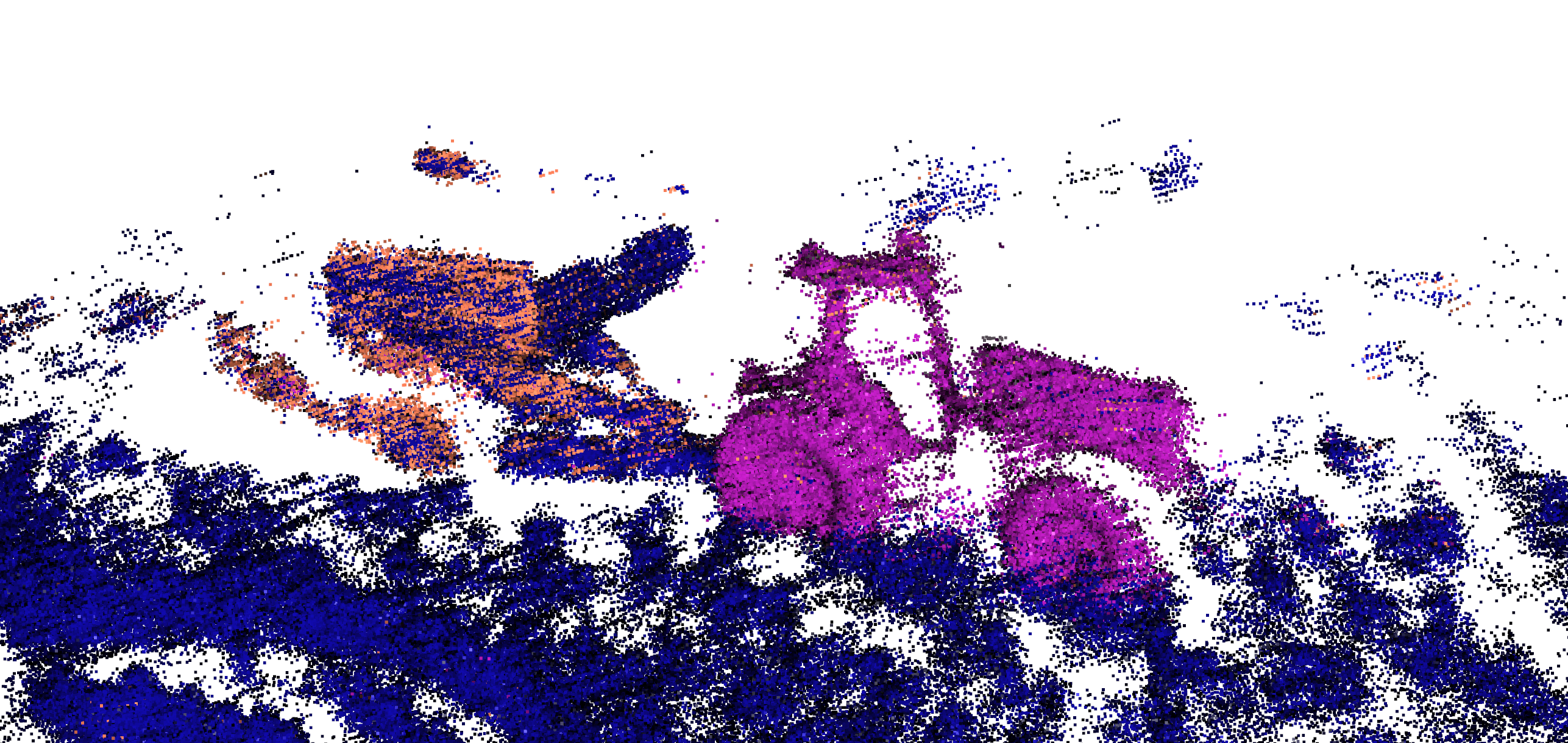}
        \end{subfigure} &
        \begin{subfigure}{0.3\textwidth}
            \centering
            \includegraphics[width=\textwidth]{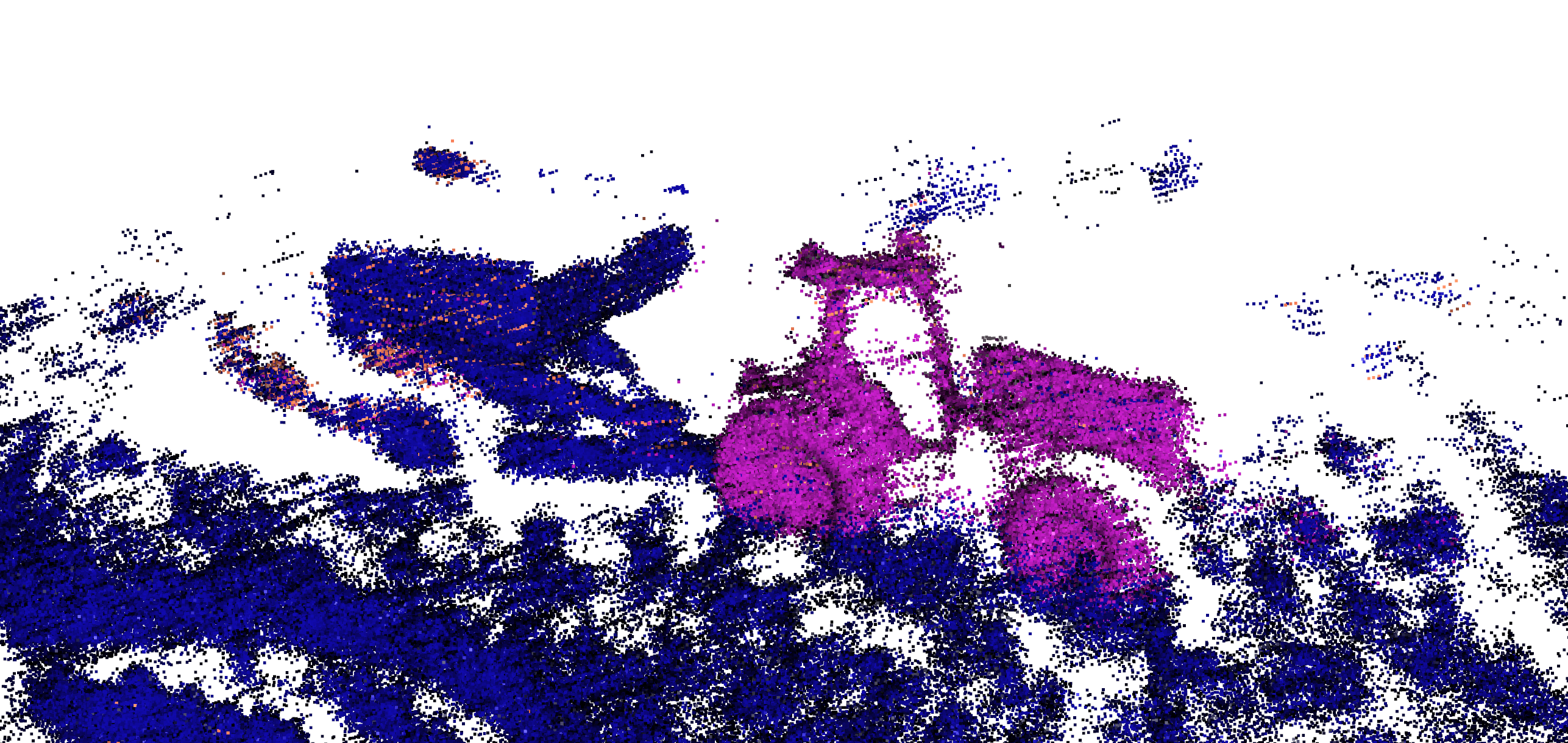}
        \end{subfigure} \\
        
        \begin{subfigure}{0.3\textwidth}
            \centering
            \includegraphics[width=\textwidth]{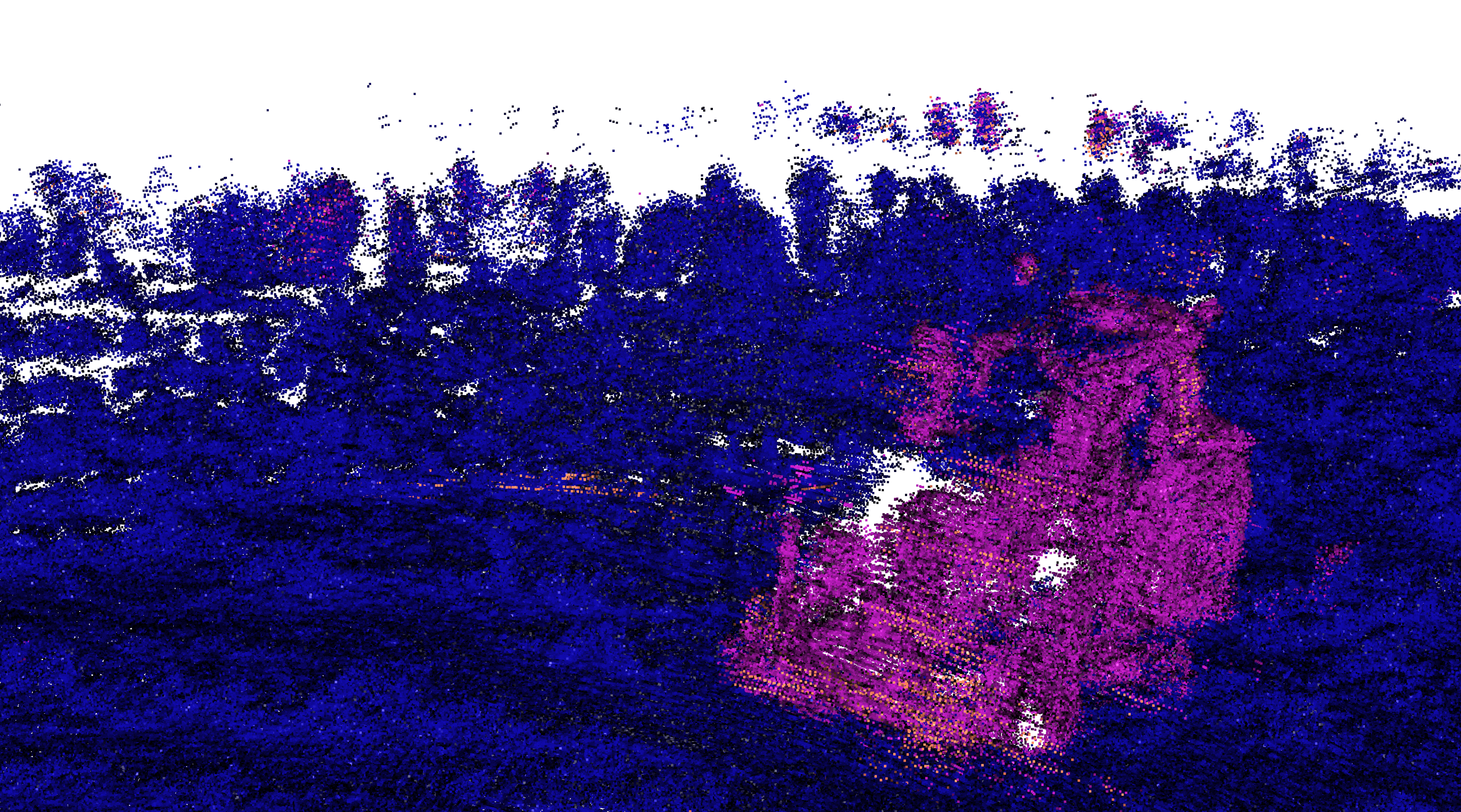}
            \caption{OACNN Baseline}
        \end{subfigure} &
        \begin{subfigure}{0.3\textwidth}
            \centering
            \includegraphics[width=\textwidth]{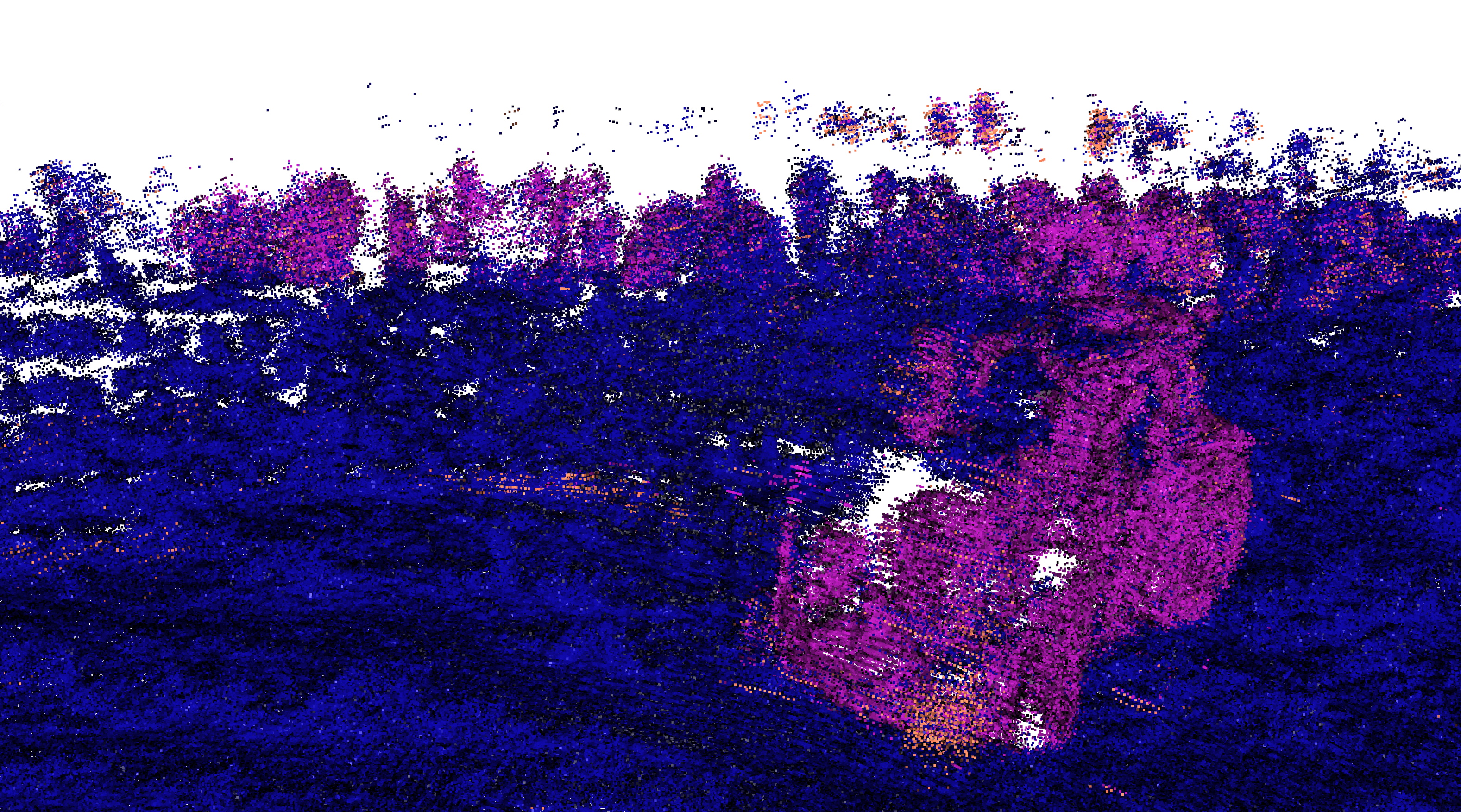}
            \caption{OACNN 10k synth}
        \end{subfigure} &
        \begin{subfigure}{0.3\textwidth}
            \centering
            \includegraphics[width=\textwidth]{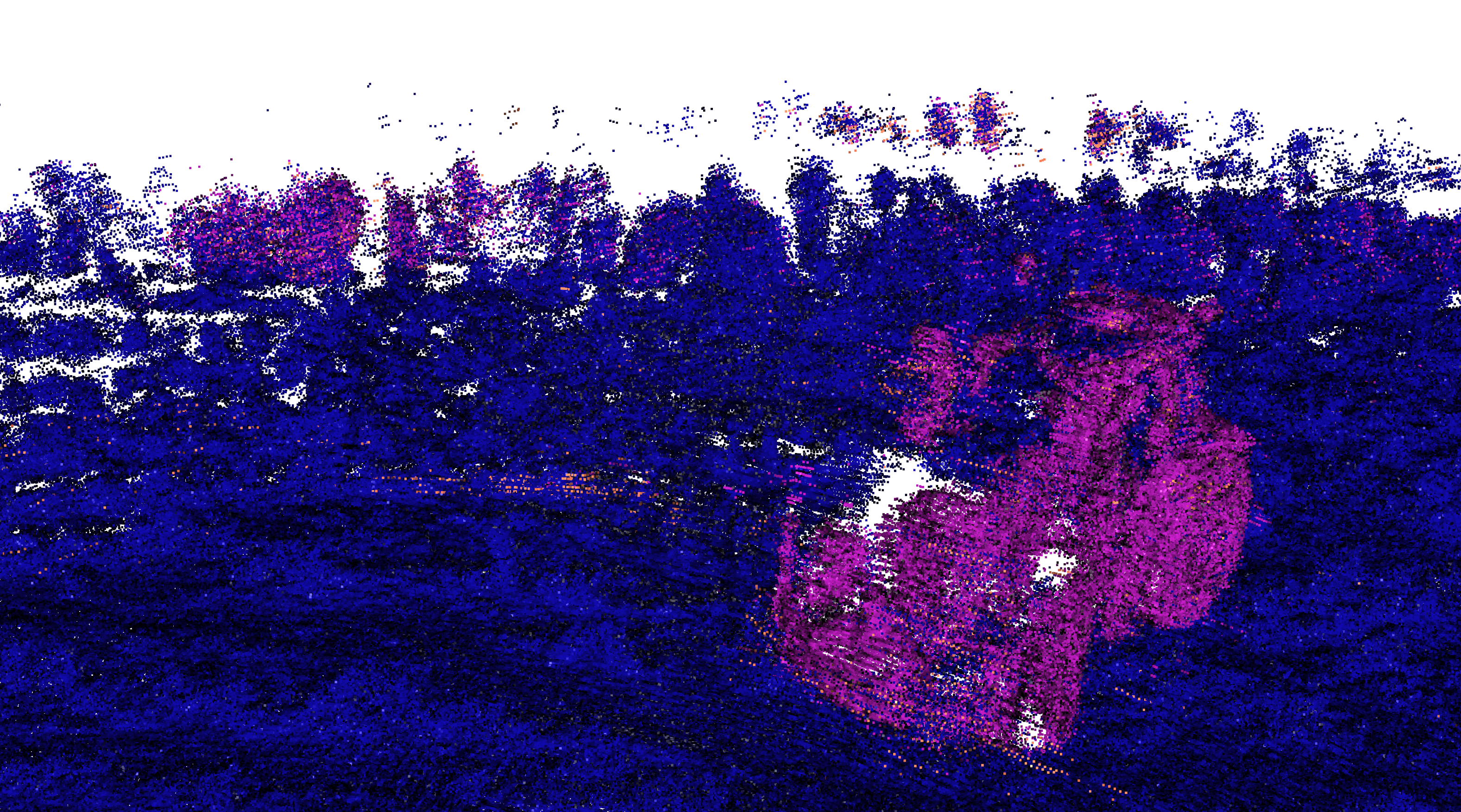}
            \caption{OACNN 65k synth}
        \end{subfigure}
    \end{tabular}
    \caption{\legendsquare{ground} other, \legendsquare{tractor} tractor, and \legendsquare{combine} combine harvester Top row: tractor with large trailer, second row: tractor with small trailer, Third row tractor with tall grass backdrop.}
    \label{fig:qual_comp}
\end{figure*}

Selection of combined point clouds with predictions from multiple models are shown in \ref{fig:qual_comp}. Notice how the baseline model can produce better prediction in the presence of tall grass, due to tall grass not being included in the data simulation.



\end{document}